\documentclass{article}

 \usepackage[preprint]{neurips_2026}

\usepackage[utf8]{inputenc} 
\usepackage[T1]{fontenc}    
\usepackage{hyperref}       
\usepackage{url}            
\usepackage{booktabs}       
\usepackage{amsfonts}       
\usepackage{nicefrac}       
\usepackage{microtype}      
\usepackage{xcolor}         
\usepackage{siunitx}
\usepackage{amsmath}
\usepackage{booktabs}
\usepackage{multirow}
\usepackage{graphicx}
\usepackage{subcaption}

\usepackage{amssymb}
\usepackage{algorithm}
\usepackage{algpseudocode}
\usepackage[normalem]{ulem}
\usepackage{etoolbox}

\title{Neuromorphic Reinforcement Learning for Quadruped Locomotion Control on Uneven Terrain}

\author{%
  Zhuangyu Han, Abhronil Sengupta \\
  School of Electrical Engineering and Computer Science\\
  Penn State University\\
  University Park, PA 16802 \\
  \texttt{zfh5141@psu.edu, sengupta@psu.edu} \\
}

\begin{document}

\maketitle

\begin{abstract}
Reinforcement learning (RL) has enabled robust quadruped locomotion over complex terrain, but most learned controllers are trained offline with backpropagation in massively parallel simulation and deployed as fixed policies, limiting adaptation to terrain variation, payload changes, actuator wear, and other real-world conditions under onboard power constraints. 
Local learning provides a potential path toward energy-aware on-robot adaptation by replacing global backpropagation graphs with updates driven by local neural states, making the learning rule more compatible with neuromorphic and in-memory computing substrates. 
This work proposes an equilibrium-propagation (EP)-based proximal policy optimization (PPO) framework for uneven-terrain quadruped locomotion. 
The controller combines a bio-inspired central pattern generator (CPG) policy with a residual postural adjustment policy, while replacing conventional backpropagation-trained policy and value networks with EP-enabled local learning. 
To train stochastic continuous-control policies with EP, we derive an EP-compatible PPO output-nudging signal and introduce a two-sided ratio clipping mechanism that stabilizes policy updates during relaxation. 
Experiments on a 12-DoF A1 quadruped show that the proposed controller achieves stable policy convergence in a two-stage uneven terrain locomotion task. Its locomotion performance is comparable to a backpropagation-trained PPO baseline in success rate, velocity tracking, actuator power, and body stability, while improving GPU memory efficiency by 4.3\(\times\) compared with backpropagation through time (BPTT). 
These results suggest that local equilibrium-based learning can support high-dimensional embodied locomotion and provide an algorithmic foundation for low-power on-robot adaptation and fine-tuning.
\end{abstract}

\section{Introduction}

Quadruped locomotion has progressed from laboratory demonstrations on flat ground to robust traversal of irregular, deformable, and cluttered terrain. This progress has been driven by controllers that coordinate cyclic stance--swing leg motions to maintain stability, generate propulsion, and recover from disturbances. Model predictive control (MPC) and RL have enabled dynamic locomotion, recovery, and agile navigation across multiple quadruped platforms \cite{kim2019highly, hwangbo2019learning, hoeller2024anymal, kumar2021rma, smith2022walk}. These advances support applications in industrial inspection, logistics, disaster response, and remote operation. However, high locomotion performance alone is insufficient for long-duration autonomy: the computational energy required by the controller becomes an increasingly important constraint.

As locomotion policies become more capable, their control-circuit cost increasingly contributes to the onboard energy budget. Conventional pipelines, including optimization-heavy MPC and backpropagation-trained RL policies, require frequent high-dimensional inference and gradient propagation on CPU/GPU platforms~\cite{kim2019highly, hwangbo2019learning, han2024learning}. This burden can further increase when controllers use multiple neural modules or external perception \cite{seto2025two, miki2022learning}. The issue is especially important for small-scale robots, where controller complexity does not necessarily decrease in proportion to mechanical power consumption. Therefore, future quadruped systems require controllers that support not only efficient inference but also energy-aware adaptation after deployment.

Most RL-based quadruped policies are trained off-robot in contact-rich simulation and transferred to hardware through sim-to-real randomization, system identification, motor adaptation, or actuator modeling~\cite{hwangbo2019learning, kumar2021rma, margolis2024rapid, tan2018sim}. Although effective, this simulation--deployment--retraining loop limits adaptation when the robot encounters unseen terrain, payload changes, actuator wear, or environmental conditions. On-robot RL has been demonstrated, but it often requires prepared arenas, external supervision, or external power supplies~\cite{haarnoja2018learning, ha2020learning}. In remote environments beyond reliable infrastructure, robots must operate from constrained power sources while adapting from real-world deployment data \cite{cully2015RobotsThatCan, pmlr-v164-bloesch22a}. This motivates algorithm--hardware co-design for low-power locomotion learning, where the learning rule itself should be compatible with efficient physical implementation.

Biological locomotor systems offer useful inspiration for this goal. Animals do not command each muscle independently; instead, high-level motor centers modulate spinal circuits that generate rhythmic locomotor patterns, while sensory feedback and reflex pathways adapt these patterns during interaction with the environment~\cite{ijspeert2008central, righetti2008pattern, kimura2007biologically, manoonpong2013neural,kiehn2006LOCOMOTORCIRCUITSMAMMALIAN, kiehn2016DecodingOrganizationSpinal, grillner2020CurrentPrinciplesMotor, grillner2021CPGsLimbedLocomotion}. CPG-based robotic controllers exploit this organization by converting actuator-level control into structured rhythmic parameters such as phase, amplitude, frequency, and orientation. In RL-based locomotion, this structure can reduce the burden on the learned policy by constraining the action space toward meaningful gait patterns~\cite{ijspeert2008central, bellegarda2022cpg, lee2020LearningQuadrupedalLocomotiona, suzuki2025FootTrajectoryKey}. However, CPGs organize limb-level motion but do not by themselves solve the computational and learning-efficiency challenges of trainable neural controllers. This motivates a dynamical-systems perspective in which oscillator dynamics structure locomotion control, while neural relaxation dynamics provide a mechanism for local learning and adaptation.

This remaining challenge connects naturally to Neuro-AI, which seeks to use principles from biological intelligence to develop artificial agents with stronger robustness, adaptation, and energy efficiency. For embodied robots, this perspective shifts the focus from only accelerating inference to designing learning rules that can adapt under local information and limited onboard resources. Backpropagation remains the dominant training method for deep RL, but it relies on global credit assignment, activation storage, and backward gradient transport, which are difficult to reconcile with biological or physical learning substrates. Recent high-impact alternatives, including feedback alignment, target propagation, predictive-coding learning, and equilibrium propagation \cite{lillicrap2016RandomSynapticFeedback, lee2015DifferenceTargetPropagation, whittington2017ApproximationErrorBackpropagation, scellier2017equilibrium}, show that multilayer network learning can be approximated or implemented through local activity, target states, or dynamical perturbations rather than exact backpropagation.

Among these approaches, EP is especially relevant because both inference and learning are expressed through network relaxation dynamics. EP trains energy-based networks by comparing nearby equilibrium states induced by small output nudges, avoiding explicit global backpropagation graphs and aligning naturally with in-memory and neuromorphic hardware dynamics~\cite{scellier2017equilibrium, laborieux2021scaling}. Recent EP variants further suggest its potential for efficient local learning \cite{laborieux2021scaling, laborieux2022holomorphic, laborieux2023improving, scellier2023energy, martin2021eqspike}, but EP has mainly been studied on supervised benchmarks. Its use in reinforcement learning remains limited, and EP-based policy and value learning has not yet been demonstrated for high-dimensional uneven-terrain quadruped locomotion.

This work addresses the gap between high-performance simulation-trained quadruped locomotion and low-power local neuromorphic learning. We propose an EP-based PPO framework for a bio-inspired quadruped controller that combines CPG-based gait generation with residual postural adjustment. The framework targets energy-efficient local on-chip learning while maintaining locomotion performance comparable to a backpropagation-trained PPO controller. The main contributions are:

\textbf{1. An EP-based PPO framework for continuous-control locomotion without backpropagation.}
We replace the backpropagation-trained policy and value networks in PPO with equilibrium-propagation networks and formulate EP-based updates for both policy learning and value regression. This provides a local-learning alternative to conventional actor–critic training and establishes a pathway toward neuromorphic implementation of reinforcement learning for large-DoF robotic control.

\textbf{2. An EP-compatible policy objective gradient for PPO training.}
We derive and implement an output-layer nudging signal that maps the PPO policy objective onto EP nudge phase equilibrium dynamics. To stabilize policy updates, we introduce a two-sided PPO ratio clip objective and an adjusted gradient-scaling rule for the EP nudge phase. This design enables EP networks to train a continuous-control policy while avoiding direct backpropagation through the policy network.

\textbf{3. A bio-inspired EP controller combining CPG-based gait generation and residual postural adjustment.}
We integrate EP policy networks into a two-branch quadruped controller composed of a CPG policy and a residual-angle policy. The CPG branch generates structured rhythmic foot trajectories, while the residual branch corrects joint targets for uneven-terrain traversal. This architecture connects biological locomotion organization with local neuromorphic learning.

We evaluate the proposed method on a Unitree A1 quadruped model over randomized uneven terrain and compare it with a backpropagation-trained PPO controller under the same task setting. The EP-based controller achieves a comparable success rate, velocity tracking, roll/pitch angular-velocity stability, and actuator power consumption, demonstrating that local EP learning can support large-DoF locomotion beyond simple neuromorphic control benchmarks.

\section{Related Works}
Biological locomotion relies on hierarchical motor organization: high-level commands modulate rhythmic spinal circuits, while sensory feedback and reflex pathways adapt movement during environment interaction~\cite{kiehn2006LOCOMOTORCIRCUITSMAMMALIAN, kiehn2016DecodingOrganizationSpinal, grillner2020CurrentPrinciplesMotor, grillner2021CPGsLimbedLocomotion}. This principle has inspired CPG-based robotic controllers, where oscillatory dynamics generate coordinated periodic limb trajectories and reduce the dimensionality of locomotion control. Classic CPG studies demonstrate stability, adaptability, and sensory-entrainment benefits for legged robots~\cite{ijspeert2008central, righetti2008pattern, kimura2007biologically, manoonpong2013neural}. In RL-based locomotion, CPGs constrain policy search toward structured rhythmic movements instead of unconstrained joint torques or joint targets. Prior work has combined CPGs with RL for hexapod and quadruped locomotion, including PPO-based CPG-RL, visually guided CPG locomotion, residual postural policies for uneven terrain, and foot-trajectory studies~\cite{ouyang2021adaptive, bellegarda2022cpg, bellegarda2024visual, seto2025two, suzuki2025FootTrajectoryKey}. These studies show that CPG architectures provide useful locomotion priors, but their trainable components are still generally optimized by backpropagation, limiting direct deployment on local-learning neuromorphic substrates.

Local learning algorithms provide a more biologically plausible approach for resource-constrained learning. Feedback alignment relaxes the weight-transport requirement of backpropagation, target propagation replaces layerwise gradients with local target states, predictive-coding networks approximate backpropagation through iterative inference and local Hebbian updates, and EP formulates learning as the contrast between free and weakly nudged equilibrium states~\cite{lillicrap2016RandomSynapticFeedback, lee2015DifferenceTargetPropagation, whittington2017ApproximationErrorBackpropagation, scellier2017equilibrium}. These methods support the Neuro-AI goal of adaptive, biologically inspired learning systems, but most have been evaluated primarily on supervised benchmarks rather than contact-rich robot learning.

Recent EP developments have improved scalability and hardware relevance through second nudging phases, holomorphic formulations, Jacobian homeostasis, asynchronous relaxation, analog-computing analyses, and spike-driven implementations~\cite{laborieux2021scaling,laborieux2022holomorphic,laborieux2023improving,scellier2023energy,martin2021eqspike}. Nevertheless, EP remains underexplored in reinforcement learning. Prior work has considered EP in actor--critic settings \cite{kubo2022CombiningBackpropagationEquilibrium}, but EP-based policy and value learning has not been demonstrated for high-dimensional legged locomotion over uneven terrain. In contrast, this work integrates EP networks into PPO, derives an EP-compatible continuous-control policy nudging signal, and evaluates the resulting CPG--residual controller on uneven-terrain quadruped locomotion.

Together, these lines of work motivate a controller that combines the structured locomotion priors of CPG-based control with local neuromorphic learning in the trainable policy and value networks. The following section describes the proposed EP-based CPG–RES quadruped controller and its PPO training procedure.

\section{Methods}
We use a two-branch CPG–RES controller to combine structured rhythmic gait generation with learned postural correction. The controller is evaluated on a 12-DoF Unitree A1 quadruped simulated in MuJoCo. Each limb contains hip, thigh, and calf joints; we use \(i\in\{1,2,3,4\}\) to index limbs and \(j\in\{\mathrm{hip},\mathrm{thigh},\mathrm{calf}\}\) to index joints within a limb. 

\begin{figure}[h]
    \centering
    \includegraphics[width=1.0\textwidth]{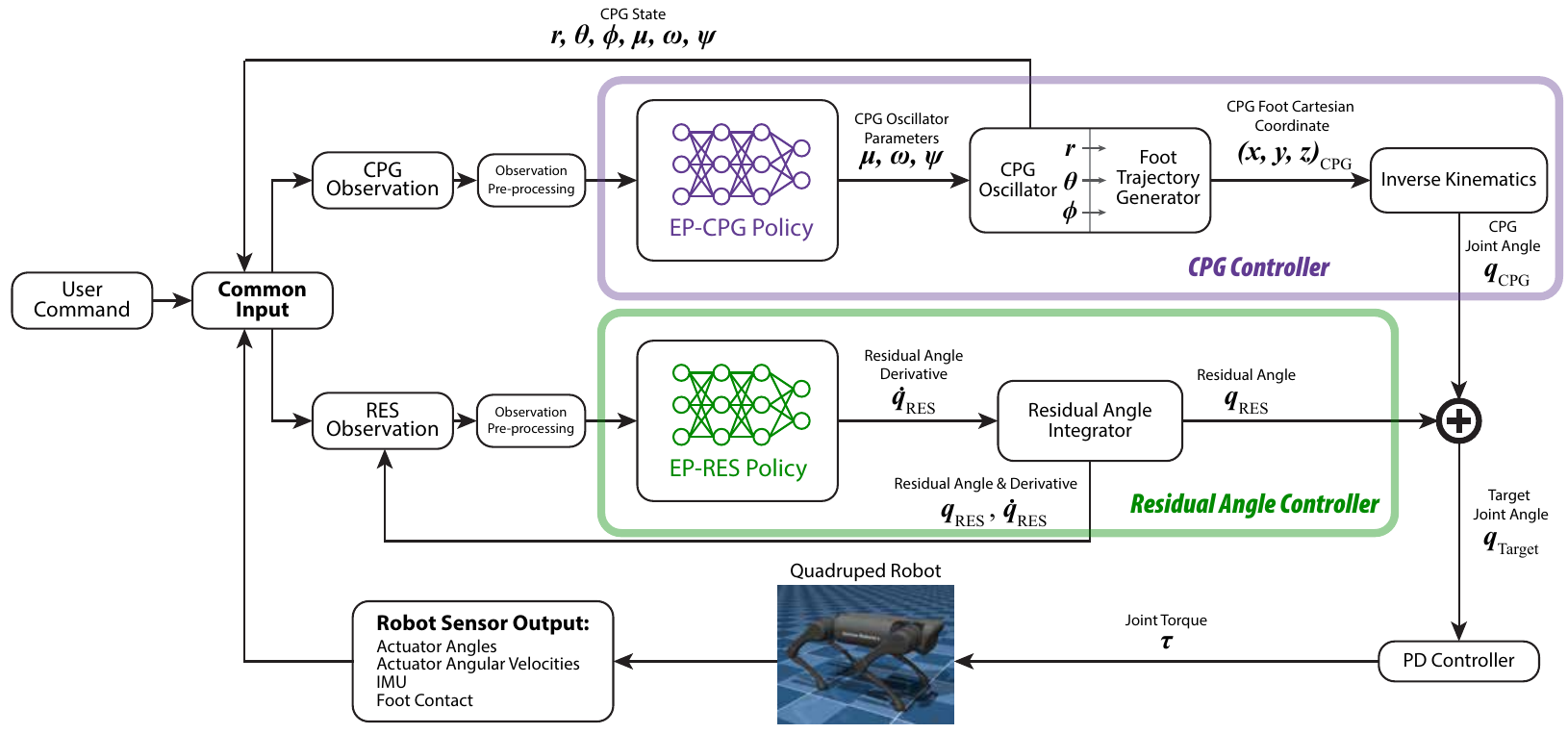}
    \caption{The control architecture of the EP-based quadruped locomotion control. This schematic shows the stage-2 uneven terrain control iteration. The training process is not shown here. The overall control architecture follows \cite{seto2025two} with minor modifications.}
    \label{ControlArchitecture}
\end{figure}

The reinforcement learning task of this paper is to drive a quadruped robot to walk robustly on uneven terrain. Several RL-based training paradigms have achieved outstanding locomotion stability and agility on various types of quadruped platforms \cite{kumar2021rma, hwangbo2019learning, margolis2024rapid}. The control architecture in this paper follows the CPG-RES two-stage learning paradigm in \cite{seto2025two}, where the core CPG-RL controller is from \cite{bellegarda2022cpg}. Fig. \ref{ControlArchitecture} shows a schematic of the control architecture. This architecture contains two RL policy networks: the CPG policy network and the RES policy network. At each RL control timestep, these two policy networks map the observation vector to two action vectors: the CPG oscillator parameter vector and the residual angle derivative vector, respectively. For the CPG branch, the CPG oscillator parameter vector is fed into the CPG oscillator. At each low-level time step, the CPG oscillator provides three CPG states \((r_i, \theta_i, \phi_i)\). The \((r_i, \theta_i, \phi_i)\) controls a foot trajectory generator, which outputs CPG foot Cartesian coordinates. The CPG foot Cartesian coordinates are converted into CPG joint angles through inverse kinematics. For the residual angle branch, integrating the residual angle derivatives gives the residual angles. The residual angles are added to the CPG joint angles, and the resulting target joint angles are converted into the joint torques through a PD controller. Joint torque is the lowest-level command for controlling a robot in both simulation and the real world.

\subsection{EP-based CPG--RES locomotion controller}

The details of the CPG oscillator, foot trajectory generator, and the residual angle integration are reported in Appendix~\ref{sec:control_A}. In brief, after the CPG controller and the residual angle controller output the CPG joint angle and the residual angle, respectively, the overall target joint angle is the sum of their outputs:
\begin{align}
    q_{j, i, \mathrm{target}} = q_{j, i, \mathrm{CPG}} + q_{j, i, \mathrm{RES}} \quad \forall j,i
\end{align}
Here \(q_{j, i, \mathrm{CPG}}\) is the CPG joint angle and \(q_{j, i,\mathrm{RES}}(t)\) is the residual angle. The PD controller has the following form:
\begin{align}
\tau_{j, i} =  \mathbf{Clamp}(-K_p(q_{j, i} - q_{j, i,\mathrm{target}}) - K_d\,\dot{q}_{j, i}, -\mathbf{TorqueLimit},\mathbf{TorqueLimit})  \quad \forall j,i
\end{align}
Here, \(\tau_{j, i}\), \(q_{j, i}\), \(q_{j, i,\mathrm{target}}\), and \(\dot{q}_{j, i}\) are joint torque to be applied, current joint angle, target joint angle, current angular velocity of joint \(j\) at limb \(i\), respectively. Two PD hyper-parameters are \(K_p = \qty{100}{\newton\meter}\) and \(K_d = \qty{2}{\newton\meter\second}\). Also, all the calculated joint torques are clamped to the actuator torque range of \([-\mathbf{TorqueLimit},\mathbf{TorqueLimit}]\) before being applied to the MuJoCo environment.

The learned CPG and RES policies run at 100 Hz, while the CPG oscillator, residual-angle integrator, PD controller, and MuJoCo simulation run at 1000 Hz. Thus, each high-level policy action is held for 10 low-level control steps. The CPG and RES policies receive 63-D and 87-D observations, respectively, and each outputs a 12-D action vector. Details about the robot parameters, oscillator dynamics, foot trajectory generator, residual integrator dynamics, observation/action definitions, reward function, and hyper-parameters are given in Appendix~\ref{sec:control_A} and \ref{sec:hyper_parameter_list}. Readers are also directed to \cite{seto2025two} for the original proposal of the CPG--RES architecture.

\subsection{EP-PPO for continuous-control learning}

We use PPO \cite{schulman2017ProximalPolicyOptimizationb} as the RL framework, but replace the backpropagation-trained policy and value networks with EP networks. The CPG policy, RES policy, and value function are all trained through equilibrium relaxation rather than direct backpropagation. 

First, assume there is only one environment. In the PPO rollout, at each control step \(t\), the following data tuple will be stored in the buffer
\[
(s_t, a_t, r_t, \mathrm{log}(\pi_{\mathrm{rollout}}(a_t|s_t)), d_t, \mathrm{fall}_t, \mathbf{Value}(s_t), \mathbf{Value}(s_{t+1}))
\]
Here, \(s_t, a_t\), and \( r_t\) denote the observation vector, action vector, and scalar reward. \(\pi_{\mathrm{rollout}}(a_t|s_t)\) is the action probability calculated in the rollout before any policy update, and ``rollout'' is used to distinguish the before-policy-update probability and after-policy-update probability. \(d_t\) is the episode ending signal (see Appendix~\ref{sec:control_A} for episode ending conditions), and \(\mathrm{fall}_t\) is the robot falling signal. \(\mathbf{Value}(s_t), \mathbf{Value}(s_{t+1})\) are the value function of \(s_t\) and \(s_{t+1}\).
In parallel simulation framework considered herein, there are \(N\) parallel environments, and, in a single rollout, each environment forwards \(T\) control steps. A total of \(N\cdot T\) data tuples will be collected in the rollout phase. In the neural network update phase, the data tuples stored in the buffer are processed. \(\forall t\), first, \(\mathbf{Return}(s_t)
\) and \(\mathbf{Advantage}(s_t, a_t)
\) are calculated through generalized advantage estimator (GAE) \cite{schulman2015HighDimensionalContinuousControl}, which requires \(r_t, d_t, \mathrm{fall}_t\) and \(\mathbf{Value}(s_t), \mathbf{Value}(s_{t+1})\). 
Subsequently, the policy network and the value network are updated for \(K_\mathrm{epoch}\) epochs and \(\frac{|\mathcal{D}|}{|\mathcal{B}|}\) mini-batches in each epoch, where \(\mathcal{D}\) is a single rollout dataset with \(N\cdot T\) data tuples, and \(\mathcal{B}\) is a data mini-batch. Training the value network is formulated as a regression task. See details about the value loss function in Appendix~\ref{sec:EP_A}.

The policy network is updated by maximizing the PPO action surrogate objective function: 
\begin{equation}
L^{\mathrm{CLIP}} = \hat{\mathbb{E}_t} [\min(r_t(\mu_t)\hat{A}_t, \mathbf{Clamp}(r_t(\mu_t), 1-\epsilon, 1+\epsilon)\hat{A}_t)]
\label{policyloss}
\end{equation}
Here, \(\hat{A}_t \) is \( \mathbf{Advantage}(s_t, a_t)\), and \(\epsilon\) is the PPO ratio clip coefficient. \(r_t(\mu_t) = \frac{\pi(a_t|s_t)}{\pi_{\mathrm{rollout}}(a_t|s_t)}\) is the probability ratio. \(\pi(a_t|s_t)\) is the current-policy probability of sampling the rollout action \(a_t \) given observation \(s_t\). 
If \(\mu_{t,i}\) is the \(i\)-th dimension of the policy output action mean in sample \(t\), given the PPO assumption \(r_t(\mu_{t}) \approx 1\) in the update process, we have the approximate PPO objective gradient w.r.t. the policy network output \(\mu_{t,i}\):
\begin{align}
\label{clip_obj_grad}
\frac{\partial L^{\mathrm{CLIP}}}{\partial \mu_{t,i}} &= \frac{1}{|\mathcal{B}|}
    \begin{cases}
      \mathbf{1}_{[0, 1+\epsilon
      )}(r_t(\mu_{t}))\cdot r_t(\mu_{t})\frac{(a_{t,i} - \mu_{t,i})}{\sigma_i^2}\hat{A}_t & \text{if \(\hat{A}_t \geq 0\)}\\
      \mathbf{1}_{(1-\epsilon,
      \infty)}(r_t(\mu_{t}))\cdot r_t(\mu_{t})\frac{(a_{t,i} - \mu_{t,i})}{\sigma_i^2}\hat{A}_t & \text{if \(\hat{A}_t < 0\)}
    \end{cases} \\
    &\approx \frac{1}{|\mathcal{B}|}
    \begin{cases}
      \mathbf{1}_{[0, 1+\epsilon
      )}(r_t(\mu_{t}))\cdot \frac{(a_{t,i} - \mu_{t,i})}{\sigma_i^2}\hat{A}_t & \text{if \(\hat{A}_t \geq 0\)}\\
      \mathbf{1}_{(1-\epsilon,
      \infty)}(r_t(\mu_{t}))\cdot \frac{(a_{t,i} - \mu_{t,i})}{\sigma_i^2}\hat{A}_t & \text{if \(\hat{A}_t < 0\)}
    \end{cases} \notag 
\end{align}
Here, the terms \(\mathbf{1}_{[0, 1+\epsilon)}(r_t(\mu_{t}))\) and \(\mathbf{1}_{(1-\epsilon,\infty)}(r_t(\mu_{t}))\) can be regarded as gradient masks. It sets the continuous part of the gradient \(\frac{(a_{t,i} - \mu_{t,i})}{\sigma_i^2}\hat{A}_t\) to zero when the probability ratio is beyond the gradient enabling range (positive advantage: \([0, 1+\epsilon)\), negative advantage: \((1-\epsilon,\infty)\)). For the full derivation of the PPO objective gradient, see Appendix~\ref{sec:EP_A}.

Beyond the PPO objective gradient, it is important to understand how the EP network optimizes a given objective function. The EP network is a dynamic system, where all the neuron states evolve according to the negative energy gradient. The energy function of EP has the following form when the task is to minimize a loss function \(L\):
\begin{equation}
    E(\xi) = \frac{1}{2}\sum_i \xi_i^2 - \sum_{i, j}w_{ij}\rho(\xi_i) \rho(\xi_j) - \sum_i b_i \rho(\xi_i) - \beta L(\xi_{out}, y)
\label{EPenergy}
\end{equation}
Here, \(\xi_i\) is the state of neuron \(i\). \(\rho(\cdot)\) is the activation function. \(w_{ij}\) and \(b_i\) are the weights and biases. \(L(\xi_{out}, y)\) is the loss function. \(\xi_{out}\) is the output layer neuron's state vector. \(\beta\) is the EP nudge coefficient. \(\beta\) determines whether the loss function influences the neuron states. Note that the \(\beta\) sign convention is different than the original formulation in \cite{scellier2017equilibrium}. The EP network converges to an equilibrium following the dynamics. For neuron \(i\) in the output layer \(\xi_{out,i}\), the time derivative is:
\begin{equation}
    \label{neurons_state_update}
    \frac{d \xi_{out,i}}{d[\mathrm{time}]} = -\frac{\partial E}{\partial \xi_{out,i}} = - \xi_{out,i} + \rho'(\xi_{out,i})\sum_{j}w_{out,ij} \rho(\xi_{j}) + b_{out,i} \rho'(\xi_{out,i}) + \beta \frac{\partial L(\xi_{out})}{\partial \xi_{out,i}}
\end{equation}
For three-phase EP, the network first reaches a free equilibrium at \(\beta = 0\). The free phase equilibrium neuron state is denoted by \(\xi^*\). Then the EP network reaches two nudge equilibrium states at \(\beta = \beta_{\mathrm{ep}}\) and \(\beta = -\beta_{\mathrm{ep}}\). The initial state value for the nudge phase relaxation is \(\xi^*\). The gradient of the loss function w.r.t. the EP network parameters \(w_{ij}\) (a weight matrix element) can be approximated by using the two nudge equilibrium states, \(\forall i,j\)
\begin{equation}
\label{ep_grad}
    \frac{\partial L(\xi^*_{out}, y)}{\partial w_{ij}}
    \approx \frac{1}{2\beta_{\mathrm{ep}}}[\rho(\xi_i^{+\beta_{\mathrm{ep}}}) \rho(\xi_j^{+\beta_{\mathrm{ep}}}) - \rho(\xi_i^{-\beta_{\mathrm{ep}}}) \rho(\xi_j^{-\beta_{\mathrm{ep}}})]
\end{equation}
Here, \(\xi_i^{+\beta_{\mathrm{ep}}}\) and \(\xi_i^{-\beta_{\mathrm{ep}}}\) are the equilibrium states at of the positive and negative nudge phases. For a bias parameter \(b_{i}\), the form of loss gradient is similar:
\begin{equation}
\label{ep_grad_bias}
    \frac{\partial L(\xi^*_{out}, y)}{\partial b_{i}}
    \approx \frac{1}{2\beta_{\mathrm{ep}}}[\rho(\xi_i^{+\beta_{\mathrm{ep}}}) - \rho(\xi_i^{-\beta_{\mathrm{ep}}})]
\end{equation}
It can be summarized that, by applying \(\beta\frac{\partial L(\xi_{out}, y)}{\partial \xi_{out}}\) at the output layer neurons in the nudge phases, the EP learning mechanism can calculate the loss gradient w.r.t. the network parameters through equilibrium differentiation.
Therefore, to optimize the PPO objective, we can apply the PPO objective gradient w.r.t. the network output \(\frac{\partial L^{\mathrm{CLIP}}}{\partial \mu_{t,i}}\) to the output layer neurons in the EP network. Note that there is no neuron state dynamics or free-nudge relaxation in the traditional backpropagation-trained feed-forward network -- there is only a single output layer activation \(\mu_{t}\) without any temporal dynamics. The \(\mu_{t,i}\) is the policy network output dimension \(i\), and is equivalent to \(\xi^*_{t,out,i}\) in EP, which is the output neuron state \(i\) in free phase equilibrium when the policy input observation is \(s_t\).

By default, the EP network wants to minimize a loss function \(L(\xi_{out}, y)\). However, here the PPO algorithm requires its objective to be maximized. Therefore, we can apply the negative of the PPO objective gradient w.r.t. the network output, shown in Equation \ref{clip_obj_grad}, directly to the EP output layer, which yields an output layer dynamics: 
\begin{align}
    \label{EP_out_dynamics_unmodified}
    \frac{d\xi_{t,out,i}}{d[\mathrm{time}]} = &
    -\xi_{t,out,i} 
    + \rho'(\xi_{t,out,i})\sum_{j}w_{out,ij} \rho(\xi_{t,j}) + b_{out,i} \rho'(\xi_{t,out,i})\\
    &- \beta\cdot \frac{1}{|\mathcal{B}|}
    \begin{cases}
      \mathbf{1}_{[0, 1+\epsilon
      )}(r_{t,\mathrm{nudging}}(\xi_{t,out}))\cdot \frac{(a_{t,i} - \xi_{t,out,i})}{\sigma_i^2}\hat{A}_t & \text{if \(\hat{A}_t \geq 0\)}\\
      \mathbf{1}_{(1-\epsilon,
      \infty)}(r_{t,\mathrm{nudging}}(\xi_{t,out}))\cdot \frac{(a_{t,i} - \xi_{t,out,i})}{\sigma_i^2}\hat{A}_t & \text{if \(\hat{A}_t < 0\)}
    \end{cases} \notag 
\end{align}
Here \(r_{t,\mathrm{nudging}}(\xi_{t,out})\) is the nudging probability ratio, which is dedicated to the nudge phase per-relaxation-iteration probability ratio calculation:
\begin{align}
\label{nudging_ratio}
    r_{t,\mathrm{nudging}}(\xi_{t,out}) &= \frac{\pi_{\mathrm{nudging}}(a_t|s_t)}{\pi_{\mathrm{rollout}}(a_t|s_t)}\\
    &=\frac{1}{\pi_{\mathrm{rollout}}(a_t|s_t)} \prod_i^{D_{\mathrm{action}}}\left[\frac{1}{\sqrt{2\pi\sigma_i^2}}\mathrm{exp}\left(-\frac{(a_{t,i} - \xi_{t,out,i})^2}{2\sigma_i^2}\right)\right] \notag 
\end{align}
Here \(\pi_{\mathrm{nudging}}(a_t|s_t)\) is the action probability in the nudge relaxation as a function of \(\xi_{t,out,i}\). \(D_{\mathrm{action}}\) is the dimensionality of action space. The reason for using the notation \(r_{t,\mathrm{nudging}}(\xi_{t,out})\) instead of \(r_{t}(\xi_{t,out})\) is that the probability ratio keeps changing/oscillating in both nudge phase iterations because \(\xi_{t,out}\) keeps changing/oscillating, and this nudging probability ratio controls the gradient mask in each relaxation step. 
However, experiments show that this formulation fails to converge when using Equation \ref{EP_out_dynamics_unmodified} as the objective gradient. The per-update KL-divergence graph indicates excessively large update steps. Upon investigation of the cause of the large update steps, we conclude that, for positive-advantage samples, the positive nudge phase will drive the output neuron state toward the \textbf{farther-away-from-target} direction without bound. For a detailed discussion about the cause of the large update step using the original PPO objective gradient, see Appendix~\ref{sec:EP_A}. To constrain these large update steps, we propose the two-sided PPO ratio clip objective gradient:
\begin{equation}
\label{modified_policy_loss_grad}
\frac{\partial L^{\mathrm{TwoSidedCLIP}}}{\partial \xi_{t,out,i}} = \frac{1}{|\mathcal{B}|}
    \begin{cases}
      \mathbf{1}_{(1-\epsilon_{\mathrm{rev}}, 1+\epsilon
      )}(r_{t,\mathrm{nudging}}(\xi_{t,out}))\cdot \frac{(a_{t,i} - \xi_{t,out,i})}{\sigma_i}\hat{A}_t & \text{if \(\hat{A}_t \geq 0\)}\\
      \mathbf{1}_{(1-\epsilon,
      1+\epsilon_{\mathrm{rev}})}(r_{t,\mathrm{nudging}}(\xi_{t,out}))\cdot \frac{(a_{t,i} - \xi_{t,out,i})}{\sigma_i}\hat{A}_t & \text{if \(\hat{A}_t < 0\)}
    \end{cases} 
\end{equation}
Here, \(\epsilon_{\mathrm{rev}}\) is the reverse PPO ratio clip coefficient. The differences between the modified policy gradient and the original version are the \textbf{two-sided ratio clipping} and the \textbf{standard deviation \(\sigma_i\) in the denominator} instead of the variance \(\sigma_i^2\). \(r_{t,\mathrm{nudging}}(\xi_{t,out})\) is calculated at each nudge phase step.
The experiments show that this two-sided bounded gradient and reduced gradient scaling factor \(\frac{1}{\sigma_i}\) successfully leads to training convergence. The ablation studies regarding different values of \(\epsilon_{\mathrm{rev}}\) and \(\frac{1}{\sigma_i}\) are shown in  Appendix~\ref{sec:ablation_study}. The additional experiments in Appendix~\ref{sec:extended_experiments} show the advantage-\(r_{\mathrm{nudging}}\) relationship and explain the policy collapse when \(\epsilon_{\mathrm{rev}} \geq 1.0\).

Because EP inference requires iterative relaxation, reducing the number of relaxation steps is important for both training time and control latency. We therefore apply an inverse discrete cosine transform (iDCT)-based observation preprocessing step before the EP networks. Empirically, this preprocessing accelerates convergence to equilibrium and reduces the required inference iterations. The ablation study in Appendix~\ref{sec:ablation_study} evaluates its effect.

Full derivations of the Gaussian policy gradient and the log-std update are provided in Appendix~\ref{sec:EP_A}.

\subsection{Locomotion training paradigm}

Training follows the two-stage paradigm of the CPG--RES controller. In stage 1, only the CPG policy is trained on flat ground to learn stable forward locomotion and velocity tracking. In stage 2, the trained CPG policy is fixed, and the RES policy is trained on randomized uneven terrain to provide joint-level postural corrections. The reward combines forward-velocity tracking, lateral/yaw tracking regularization, body-stability penalties, and actuator-power regularization.
Rollouts are collected from parallel MuJoCo environments. EP-PPO algorithm pseudo-code is provided in Appendix~\ref{sec:epppo}. Detailed hyper-parameters, domain randomization ranges, terrain generation, robot initialization, observation preprocessing, and EP network architecture/setup are provided in Appendices~\ref{sec:mujoco_A}, \ref{sec:EP_A}, and \ref{sec:hyper_parameter_list}.

\section{Experiments}
The training experiments were conducted using up to two NVIDIA RTX A5000 GPUs. Each GPU has 24,564 MB of memory. The CPU memory consumption by stage-1 and stage-2 is approximately 60 GB and 200 GB, respectively. The massively parallel MuJoCo environment consumes the majority of the CPU memory.

For the uneven terrain locomotion task described in the Methods section, we use EP to train a policy network and a value network. We also use the backpropagation version of PPO to train a control policy and value network on the same locomotion task. In this section, ``BP'' represents backpropagation. The EP and BP training use the same PPO setup, rollout size, reward function, and controller architecture. The only difference is the neural network architecture and the iDCT observation preprocessing. The BP version uses the normalized observation vector without iDCT. The EP and BP training process and result can be evaluated through two main aspects: the sample efficiency and the uneven terrain locomotion performance. In Appendix~\ref{sec:ablation_study}, more details about training convergence and probability ratio variation in nudge relaxation are discussed. There is also a set of ablation studies discussing the effect of different training setups, which are provided in Appendix~\ref{sec:ablation_study}. 
The foot trajectory hyper-parameters and other environment/robot model factors are fixed in the evaluation. See Appendix~\ref{sec:mujoco_A} for details.

\subsection{Training convergence and sample efficiency}
\begin{figure}[h]
    \centering
    \includegraphics[width=1.0\textwidth]{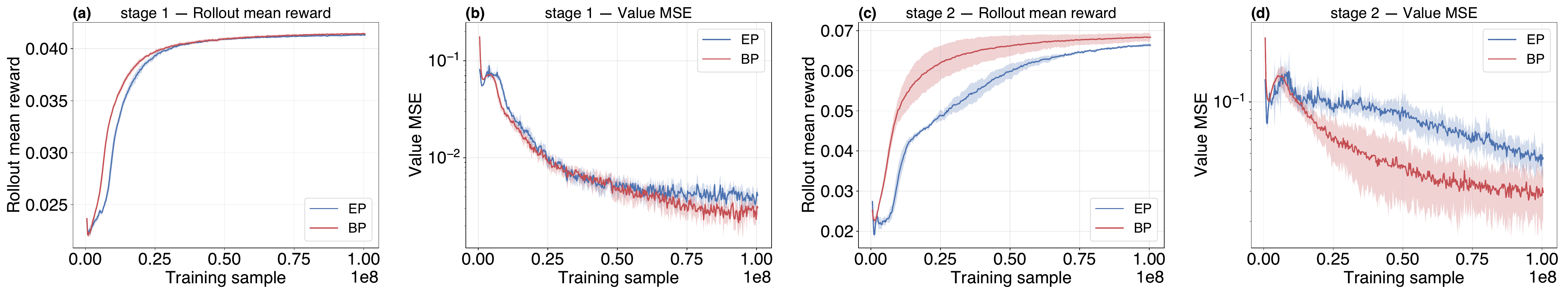}
    \caption{The convergence process of the locomotion training of BP and EP for stage-1 and stage-2. (a) and (b) shows the stage-1 rollout mean reward and MSE error between the value network output and the rollout return. (c) and (d) shows the same metric for the stage-2 training.}
    \label{convergence_grid}
\end{figure}

Sample efficiency is an important criterion for RL algorithms. BP version and EP version of PPO are used in both stage-1 and stage-2 training. Both stages are trained for \(10^8\) samples. For both EP and BP, we trained 5 independent run versions through random seeding. Note that before training an RES policy network in stage-2, a CPG policy must be trained in advance and loaded into the stage-2 training program. Each version contains a CPG policy and an RES policy that are trained sequentially, i.e., the RES policy of version 1 is trained based on the CPG policy of version 1, and the RES policy of version 2 is trained based on the CPG policy of version 2, and so on. Fig. \ref{convergence_grid} depicts the variation of rollout average per-step reward versus the number of training samples. The center line and the shaded region indicate the mean \(\pm\) standard deviation over the 5 versions. Fig. \ref{convergence_grid} (a),(c) shows that, in both stages, EP and BP reach similar final reward levels, and EP can stably improve the training reward without policy collapse. In stage-1, EP and BP converge with similar sample complexity. Also, through Fig. \ref{convergence_grid} (b),(d), the value MSE versus training samples indicates that both EP and BP value networks can continuously minimize the value loss. The value MSE is defined as the mean squared error between the value function output and the rollout return.
\begin{figure}[h]
    \centering    \includegraphics[width=1.0\textwidth]{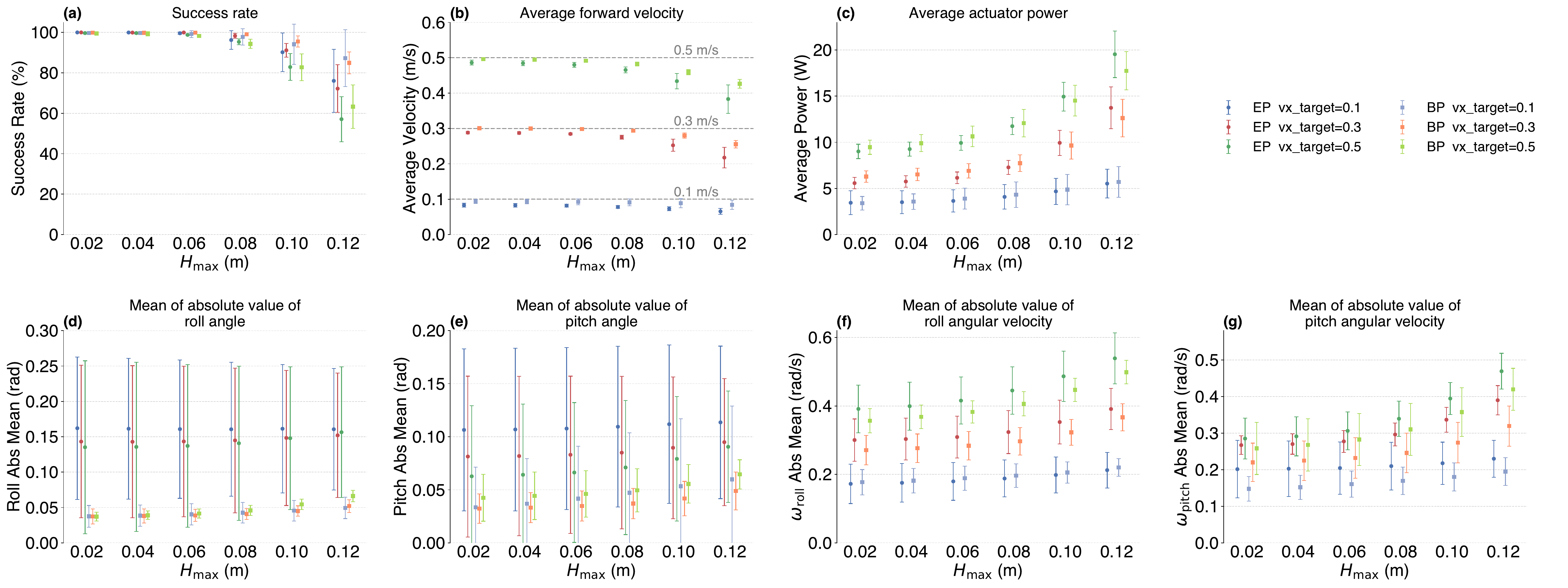}
    \caption{Locomotion performance measured in the walking test. From (a) to (g), the indicators are success rate, average forward velocity, average actuator total power, average absolute roll/pitch angle, and average absolute roll/pitch angular velocity. The center line and the error bar indicate the mean \(\pm\) standard deviation over the 5 versions.}
    \label{locomotion_performance}
\end{figure}
\subsection{Locomotion performance on uneven terrain}
We use a walking test, similar to the 5-meter walking test in \cite{seto2025two}, to measure the capability of the control policy to maintain the robot's stability when traveling across obstacles on uneven terrain. On the uneven terrain, both CPG and RES policies are enabled. The following locomotion performance indicators are recorded: success rate, average forward velocity, average actuator total power, average absolute roll/pitch angle, and average absolute roll/pitch angular velocity. The success rate is defined as follows. For each target velocity, there is a target traveling distance. If the robot can reach this target traveling distance within 30 seconds, this episode is marked as a success. Both EP and BP-trained policies are tested. For each algorithm, we test three different target velocities, and each target velocity is tested on uneven terrain with six different \(\mathbf{H_{max}}\), which indicates the hardness of the uneven terrain. A statistic over 500 episodes of walking test is conducted for each combination of conditions for each training version (note that there are 5 training versions): algorithm, target velocity, and \(\mathbf{H_{max}}\). For the details about the setup of the walking test, see Appendix~\ref{sec:extended_experiments}. The final statistics is the mean (over 5 versions) of the mean (over 500 episodes) and the standard deviation (over 5 versions) of the mean (over 500 episodes). Note that the means (over 500 episodes) of the six indicators, average forward velocity, average actuator total power, average absolute roll/pitch angle, and average absolute roll/pitch angular velocity, are weighted means. Fig. \ref{locomotion_performance} shows the over-version mean (center mark) and the over-version standard deviation (half length of the error bar) of the seven performance indicators under different testing conditions.
It can be observed that the EP and BP algorithms achieve a similar level of success rate, average forward velocity, average actuator power, and mean absolute roll/pitch angular velocity. In Appendix~\ref{sec:extended_experiments}, more details about the experiment setup and result statistics are reported.

\subsubsection{GPU memory consumption}
The gradient of the convergent RNN can also be estimated by using global BPTT \cite{werbos1990BackpropagationTimeWhat}. BPTT-based gradient estimation is typically used as a baseline in EP algorithms research to quantify computational benefit \cite{laborieux2021scaling}. 
We measured the GPU memory consumption of BPTT-based and EP local learning rule-based gradient estimation. In the experiment, the training mini-batch size is 32768. During stage-1 training, BPTT algorithm consumes 13486 MB of GPU memory, while EP algorithm consumes 3110 MB for the same task. Thus, it can be concluded that EP is approximately \(4.3\times\) more memory efficient compared with BPTT due to local computations. 

\subsection{Ablation studies}
Summary of the various ablation studies reported in Appendix~\ref{sec:ablation_study} is listed here. The training convergence of different reverse PPO ratio clip \(\epsilon_{\mathrm{rev}}\) coefficients is measured. \(1/\sigma_i\) and \(1/\sigma_i^2\) scalings lead to different training stability, which is compared. The failure of convergence as a result of the static gradient mask is displayed. The results verify the effectiveness of the proposed EP-specific modifications on top of the original PPO objective gradient (Equation \ref{clip_obj_grad}). The effect of iDCT observation preprocessing in reducing the steps to convergence in the EP relaxation is also reported.

\section{Discussion and Limitations}

This work demonstrates that EP can be integrated with PPO to train a high-dimensional quadruped locomotion controller without direct backpropagation through the policy and value networks. Combined with the CPG–RES architecture, the proposed EP-PPO controller achieves performance comparable to the BP-PPO baseline in success rate, velocity tracking, roll/pitch angular-velocity stability, and actuator power, while converging stably in both training stages without policy collapse. The EP-trained value network also steadily reduces value loss. These results suggest that local equilibrium-based learning can support contact-rich embodied control beyond low-dimensional neuromorphic benchmarks and provide an algorithmic path toward neuromorphic controllers capable of on-robot adaptation in unknown environments. More broadly, because EP updates are expressed through local equilibrium-state differences, the framework is conceptually compatible with neuromorphic and in-memory substrates, making this study an algorithmic step toward energy-aware continual locomotion learning rather than a complete hardware realization.

Several limitations remain. First, the current method inherits PPO’s limited sample efficiency: although simulation can exploit massive parallelism, real robots can only generate roughly 50–200 sequential training samples per second \cite{margolis2024rapid, chen2023LearningTorqueControl}, and unsafe exploration or repeated falls are impractical. Future work should therefore combine EP learning with more sample-efficient and safety-aware RL methods. Second, the present implementation still relies on digital computation for relaxation, rollout storage, advantage estimation, normalization, KL monitoring, and batch updates. Practical on-robot deployment will require a system-level neuromorphic design that includes both EP network dynamics and the surrounding RL infrastructure. Overall, the proposed EP-PPO framework establishes a concrete bridge between local neuromorphic learning and modern RL-based quadruped locomotion, motivating future work on sample efficiency, safety, relaxation and memory costs, and hardware-oriented EP implementations for low-power on-robot adaptation.

\section*{Acknowledgements}
The authors thank Ryosei Seto from Tohoku University for valuable discussions and guidance on reproducing the results of his previous work, which served as an important reference for this study.

Research was sponsored primarily by the Army Research Office and was accomplished under Grant Number W911NF-24-1-0127. The views and conclusions contained in this document are those of the authors and should not be interpreted as representing the official policies, either expressed or implied, of the Army Research Office or the U.S. Government. The U.S. Government is authorized to reproduce and distribute reprints for Government purposes notwithstanding any copyright notation herein.


\newpage
\appendix

\section*{Appendix}

\renewcommand{\theequation}{A-\arabic{equation}}
\setcounter{equation}{0}

\section{Control architecture}
\label{sec:control_A}

This appendix provides the implementation details of the CPG–RES controller summarized in Section 3.1, including the oscillator dynamics, foot trajectory generator, joint inverse-kinematics, and residual angle integration. This control architecture follows the two-stage learning of CPG and postural reflex proposed in \cite{seto2025two}.

\subsection{A1 robot mechanical properties}

The robot used in this work, Unitree A1, contains 4 limbs, and each limb contains 3 joints: hip, thigh, and calf. The nominal whole body weight used in the MuJoCo simulation is \(\qty{12.454}{\kilogram}\), provided by the robot manufacturer.

\subsection{CPG oscillator model and foot trajectory generator}

The Hopf oscillator is used as the core of the CPG following \cite{seto2025two}. Four independent Hopf oscillators generate periodic signals for the four feet. The dynamics of the Hopf oscillator for foot \(i\) are, \(\forall i\)
\begin{align}
& \ddot{r}_i = a(\frac{a}{4}(\mu_i - r_i) - \dot{r}_i) \\
& \dot{\theta}_i = \omega_i \\
& \dot{\phi}_i = \psi_i
\end{align}
Here, \(\mu_i, \omega_i, \psi_i\) are the CPG oscillator parameters that are provided by the CPG policy network at each RL control step. \(a = \qty{150}{\second}^{-1}\). The \(r_i, \theta_i, \phi_i\) are the CPG oscillator states. \(r_i \in \mathbb{R}\). \(\theta_i \) and \( \phi_i\) are wrapped into \([-\pi, \pi]\). \(r_i, \theta_i, \phi_i\) directly determine the CPG Cartesian coordinates of the foot \(i\) in the robot local frame in the following way, \(\forall i\)
\begin{align}
& x_{i,\mathrm{CPG}} = -d_{\mathrm{step}}(r_i-1)\mathrm{cos}(\theta_i)\mathrm{cos}(\phi_i)\\
& y_{i,\mathrm{CPG}} = -d_{\mathrm{step}}(r_i-1)\mathrm{cos}(\theta_i)\mathrm{sin}(\phi_i)\\
& z_{i,\mathrm{CPG}} = -h + (g_c \,\, \mathbf{if}\,\, \mathrm{sin}(\theta_i) > 0 \,\,\mathbf{else}\,\, g_p)\mathrm{sin}(\theta_i)
\end{align}

Here, \((x_{i,\mathrm{CPG}}, y_{i,\mathrm{CPG}}, z_{i,\mathrm{CPG}})\) are the CPG Cartesian coordinates for foot \(i\). \(d_{\mathrm{step}}, h, g_c, g_p\) are the stride length, robot body height, swing clearance, and stance penetration of the foot trajectory, respectively. The \(d_{\mathrm{step}}\) parameter is fixed at \(\qty{0.15}{\meter}\), and \(h, g_c, g_p\) are randomized in the training process. For details about the randomized trajectory, see the domain randomization section. Readers are directed to \cite{seto2025two} for more details about the foot trajectory generator. Another work comprehensively discussed the emergence of different locomotion gaits that are determined by the shape of the foot trajectory \cite{suzuki2025FootTrajectoryKey}.

The hip, thigh, and calf CPG joint angles of limb \(i\): \((q_{\mathrm{hip}, i,\mathrm{CPG}}, q_{\mathrm{thigh}, i,\mathrm{CPG}}, q_{\mathrm{calf}, i,\mathrm{CPG}})\) are determined by the robot limb inverse kinematics given \((x_{i,\mathrm{CPG}}, y_{i,\mathrm{CPG}}, z_{i,\mathrm{CPG}})\):
\begin{align}
(q_{\mathrm{hip}, i,\mathrm{CPG}}, q_{\mathrm{thigh}, i,\mathrm{CPG}}, q_{\mathrm{calf}, i,\mathrm{CPG}}) = \mathbf{InverseKinematics}(x_{i,\mathrm{CPG}}, y_{i,\mathrm{CPG}}, z_{i,\mathrm{CPG}})   \quad \forall i
\end{align}

The RES policy outputs a residual angle derivative for each actuated joint. The residual angle derivative is integrated into the joint residual angle:
\begin{align}
q_{j, i,\mathrm{RES}}(t) = \mathbf{Clamp}(q_{j, i,\mathrm{RES}}(t-1) + \dot{q}_{j, i,\mathrm{RES}}(t)\cdot \Delta t_{\mathrm{low}}, -\mathbf{ResLimit}, \mathbf{ResLimit}) \quad \forall j,i
\end{align}
Here \(q_{j, i,\mathrm{RES}}(t), \dot{q}_{j, i,\mathrm{RES}}(t)\) are the residual angle and residual angle derivative for the joint \(j\) in limb \(i\) at time step \(t\). \(\Delta t_{\mathrm{low}}\) is the low-level control timestep and will be discussed later. \(\mathbf{ResLimit}\) is the bound for residual angle output. Finally, the target joint angle is the sum of the CPG joint angle and the residual angle:
\begin{align}
    q_{j, i, \mathrm{target}} = q_{j, i, \mathrm{CPG}} + q_{j, i, \mathrm{RES}} \quad \forall j,i
\end{align}

The PD controller has the following form:
\begin{align}
\tau_{j, i} =  \mathbf{Clamp}(-K_p(q_{j, i} - q_{j, i,\mathrm{target}}) - K_d\,\dot{q}_{j, i}, -\mathbf{TorqueLimit},\mathbf{TorqueLimit})  \quad \forall j,i
\end{align}

Here, \(\tau_{j, i}\), \(q_{j, i}\), \(q_{j, i,\mathrm{target}}\), and \(\dot{q}_{j, i}\) are joint torque to be applied, current joint angle, target joint angle, current angular velocity of joint \(j\) at limb \(i\), respectively. Two PD hyper-parameters are \(K_p = \qty{100}{\newton\meter}\) and \(K_d = \qty{2}{\newton\meter\second}\). Also, all the calculated joint torques are clamped to the range of \([-\mathbf{TorqueLimit},\mathbf{TorqueLimit}]\) before being applied to the MuJoCo environment.

The RL controller (CPG and RES policy networks) inference frequency is 100 Hz, i.e., \(\Delta t_{\mathrm{RL}} = 0.01 \si{\second}\). The low-level controllers, including the PD controller, CPG oscillator dynamics, and residual angle dynamics, operate at 1000 Hz. i.e., \(\Delta t_{\mathrm{low}} = 0.001 \si{\second}\). It implies that for each pair of outputs from the CPG policy network and the RES policy network, the CPG oscillator and residual angle dynamics step forward by 10 times using the same input parameters.

\subsection{Observation and action spaces}
The definition of the dimensions in the observation space and action space is shown in Table \ref{Observations} and Table \ref{Actions}, which are identical to the observation and action spaces in \cite{seto2025two}. Note that the binary foot contact dimension is 1 when the contact force between the foot and the terrain is greater than 0.1 \(\si{\newton}\) and 0 otherwise. Also, the user command contains 3 dimensions: local x-directional target velocity, local y-directional target velocity, and yaw target angular velocity. The ``local'' here refers to the local coordinate frame fixed on the robot. In the local frame, positive x-direction is the heading forward direction, positive y-direction points to the left side, and positive z-direction is the vertically upward direction. In the training process, the local y-directional target velocity and yaw target angular velocity are constantly zero in this locomotion task, i.e., this locomotion task does not require any lateral velocity or turning of the robot. For the local x-directional target velocity, at the beginning of each episode, a \(v_{x,\mathrm{local,target}}\) will be sampled from \(\mathbf{Uniform}(0, v_{x,\mathrm{upper}})\) where \(v_{x,\mathrm{upper}}\) is the upper bound of the local x-directional target velocity.

\begin{table}[t]
\centering

\begin{minipage}[t]{0.5\linewidth}
\centering
\caption{Observation spaces of CPG policy network and RES policy network.}
\label{Observations}
\begin{tabular}{@{} >{\raggedright\arraybackslash}p{3.7cm} >{\centering\arraybackslash}p{1.2cm} >{\centering\arraybackslash}p{1.2cm} @{}}
\toprule
\multirow{2}{*}{\textbf{Observation}} & \multicolumn{2}{c}{\textbf{Number of dim.}} \\
& CPG & RES \\
\midrule
Actuator angle                  & 12 & 12 \\
Actuator angular velocity       & 12 & 12 \\
IMU (euler)                     &  2 &  2 \\
IMU (angular velocity)          &  3 &  3 \\
IMU (linear acceleration)       &  3 &  3 \\
Foot contact                    &  4 &  4 \\
\midrule
CPG state                       & 24 & 24 \\
Command                         &  3 &  3 \\
\midrule
Residual angle                  & -- & 12 \\
Residual angle derivative       & -- & 12 \\
\midrule
Total                           & 63 & 87 \\
\bottomrule
\end{tabular}
\end{minipage}
\hfill
\begin{minipage}[t]{0.45\linewidth}
\centering
\caption{Action spaces of CPG policy network and RES policy network. Limb index \(i\in\{1,2,3,4\}\).}
\label{Actions}
\begin{tabular}{@{} >{\raggedright\arraybackslash}p{1.4cm} >{\centering\arraybackslash}p{2cm} >{\centering\arraybackslash}p{2cm} @{}}
\toprule
& \textbf{Action} & \textbf{Range} \\
\midrule
\multirow{3}{*}{CPG} 
& \(\mu_i\) & [1.0, 2.0] \\
& \(\omega_i\) & [0.0, 3.0] \\
& \(\psi_i\) & [-1.5, 1.5] \\
\midrule
\multirow{3}{*}{RES} 
& \(\dot{q}_{\mathrm{hip},i,\mathrm{RES}}\)   & [-5.0, 5.0] \\
& \(\dot{q}_{\mathrm{thigh},i,\mathrm{RES}}\) & [-5.0, 5.0] \\
& \(\dot{q}_{\mathrm{calf},i,\mathrm{RES}}\)  & [-5.0, 5.0] \\
\bottomrule
\end{tabular}
\end{minipage}

\end{table}

\subsection{Reward function}
The reward function indicates the accuracy of the velocity tracking, the intensity of the robot vibration, and the locomotion energy consumption. Based on the simple reward function proposed in \cite{seto2025two, bellegarda2022cpg}, modifications are made in the x-directional linear velocity tracking term and its coefficient. The details about the reward function are shown in Table \ref{reward_function}. The closeness functions \(f_{r1}(\cdot)\) and \(f_{r2}(\cdot)\) are defined as:
\begin{equation}
    f_{r1}(x) = \mathrm{exp}\left(-\frac{x^2}{d_{r1}}\right),\quad f_{r2}(x) = \mathrm{exp}\left(-\frac{x^2}{0.25}\right)
\end{equation}
Here, \(d_{r1}\) is \(0.25\) in stage-1 and \(0.04\) in stage 2.

\begin{table}[h]
\centering
\caption{Reward Function. The terms in the RL reward function. The reward coefficient for the x-directional linear velocity tracking is different in the stages. The details are in the Reward function section.}
\label{reward_function}
\begin{tabular}{@{} >{\raggedright\arraybackslash}p{4cm} >{\centering\arraybackslash}p{2cm} >{\centering\arraybackslash}p{6.5cm} @{}}
\toprule
\textbf{Reward term} & \textbf{Coefficient} & \textbf{Description}\\
\midrule
\(f_{r1}(v_{x,\mathrm{local}} - v_{x,\mathrm{local,target}})\) & \(\alpha_{vx} \Delta t_\mathrm{RL}\) & x-directional linear velocity tracking\\
\(f_{r2}(v_{y,\mathrm{local}} - v_{y,\mathrm{local,target}})\) & \(0.75\Delta t_\mathrm{RL}\) & y-directional linear velocity tracking\\
\(f_{r2}(\omega_{\mathrm{yaw}} - \omega_{{\mathrm{yaw}},\mathrm{target}})\) & \(0.50\Delta t_\mathrm{RL}\) & yaw angular velocity tracking\\
\(-v_{z,\mathrm{local}}^2\) & \(2.00\Delta t_\mathrm{RL}\) & z-directional linear velocity velocity penalty\\
\(-(\omega_{\mathrm{roll}}^2+\omega_{\mathrm{pitch}}^2)\) & \(0.05\Delta t_\mathrm{RL}\) & roll and pitch angular velocity penalty\\
\(-\sum_{i=1}^{D_{\mathrm{action}}}\tau_i\dot q_i\) & \(0.001\Delta t_\mathrm{RL}\) & total motor power\\

\bottomrule
\end{tabular}
\end{table}

Similarly, the reward coefficient for the x-directional linear velocity tracking \(\alpha_{vx}\) is different for the two stages. \(\alpha_{vx}\) is \(3.00\) in stage-1 and \(6.00\) in stage-2.

Note that the linear velocities are the velocities of the torso in the robot frame. The \(v_{x,\mathrm{local,target}}, v_{y,\mathrm{local,target}}, \omega_{\mathrm{yaw},\mathrm{target}}\) are the 3-D user commands, and they are included in the observation vector of both stage-1 and stage-2.

\subsection{Episode setup and training paradigm}
For episodic training, the maximum episode length is 2000 RL control steps, which is equivalent to \(\qty{20}{\second}\). The episode will end early if the torso or any thigh of the robot contacts the terrain. 

The entire training paradigm can be split into two stages: stage-1 and stage-2. In stage-1, the CPG policy network is trained alone on flat ground, so that it can maintain a robust forward walking gait and to track the target velocity command. The RES policy network is not enabled in stage-1. The CPG joint angles output from the foot trajectory generators are directly converted into the joint torques without adding any residual angle. In stage-2, the CPG policy network is fixed and does inference only, and the RES policy network is trained. Each episode in stage-2 uses randomly generated uneven terrain. The trained RES policy network is expected to adjust the output from the foot trajectory generator by adding residual angles onto the CPG target joint angles. The RES policy maintains the robot's stability and overcomes the terrain obstacles. Domain randomization is used in both stage-1 and stage-2, but the randomness in these two stages is slightly different.

\newpage

\section{Simulation Environment and Episode Initialization}
\label{sec:mujoco_A}

MuJoCo platform \cite{todorov2012mujoco} is used to simulate the locomotion task. MuJoCo is configured so that its forward time step size is the same as the low-level controller time step size. i.e., \(\Delta t_{\mathrm{MuJoCo}} = \Delta t_{\mathrm{low}}\). From a programming perspective, every low-level controller step is followed by a MuJoCo forward step.

\subsection{Domain randomization}

The details of domain randomization for both stages are shown in Table \ref{Domain_randomization}. The majority of the randomized hyper-parameters align with the ones in \cite{seto2025two}. However, the frequency of random push is increased. Note that the limb mass variation ratio is multiplied by every single limb link mass at the environment resetting. i.e., for each link (Unitree A1 has a total of 12 limb links), a new random mass ratio is generated and multiplied by the link mass. Also, the random push is applied horizontally (i.e., parallel with the x-y plane in global frame) with magnitude \(\qty{0.5}{\meter/\second}\). When implementing in MuJoCo, the random push is realized by applying an external force \(F_{\mathrm{push}}\) to the robot torso for 10 \(\Delta t_{\mathrm{low}}\). \(F_{\mathrm{push}}\) satisfies
\[
|F_{\mathrm{push}}| \cdot 10 \cdot \Delta t_{\mathrm{low}} = M_{\mathrm{robot}} \cdot \qty{0.5}{\meter/\second}
\]

Here, \(|F_{\mathrm{push}}|\) is the magnitude of the pushing force, and \(M_{\mathrm{robot}}\) is the mass of the entire robot in the current episode. The sequence of the timing of random push is a Poisson train, which is approximated through sampling a Bernoulli random variable in every RL control step.

\begin{table}[h]
\centering

\caption{Domain randomization in training}
\label{Domain_randomization}
\begin{tabular}{@{} >{\raggedright\arraybackslash}p{6cm} >{\centering\arraybackslash}p{6cm} @{}}
\toprule
\textbf{Environment hyper-parameter} & \textbf{Range/Value}\\
\midrule
MuJoCo ground friction sliding coefficient & stage-1,2: \([0.5, 2.5]\) \\
Limb mass ratio & stage-1,2: \([0.5, 1.5]\) \\
Torso load (\(\si{\kilogram}\)) & stage-1,2: \([0.0, 5.0]\) \\
Robot body height \(h\) (\(\si{\meter}\)) & stage-1,2: \([0.22, 0.32]\) \\
Swing clearance \(g_c\) (\(\si{\meter}\)) & stage-1: \([0.03, 0.20]\), stage-2: \([0.15, 0.20]\) \\
Stance penetration \(g_p\) (\(\si{\meter}\)) & stage-1,2: \([0.00, 0.02]\)\\
Random push average interval (\(\si{\second}\)) & stage-1,2: \(5\)\\
\bottomrule
\end{tabular}
\end{table}

The domain randomized environment and foot trajectory hyper-parameters are fixed in the evaluation, which is shown in  Table \ref{eval_hyper_parameters}.

\begin{table}[h]
\centering
\caption{Fixed walking testing hyper-parameters}
\label{eval_hyper_parameters}
\begin{tabular}{@{} >{\raggedright\arraybackslash}p{6cm} >{\centering\arraybackslash}p{3cm} @{}}
\toprule
\textbf{Environment hyper-parameter} & \textbf{Value}\\
\midrule
MuJoCo ground friction sliding coefficient  & \(1.5\) \\
Limb mass ratio &  \(1.0\) \\
Torso load (\(\si{\kilogram}\)) &  \(0.0\) \\
Terrain box side length (\(\si{\meter}\)) & \(0.4\) \\
Robot body height \(h\) (\(\si{\meter}\)) & \(0.25\) \\
Swing clearance \(g_c\) (\(\si{\meter}\)) & \(0.1\) \\
Stance penetration \(g_p\) (\(\si{\meter}\)) & \(0.02\) \\
Random push & Disabled\\
\bottomrule
\end{tabular}
\end{table}

\subsection{Uneven terrain generation}

In stage-2, when resetting the environment at the beginning of each episode, a new uneven terrain is generated by adding geom boxes onto the flat ground. All the boxes are rectangular prisms with square bases. The shape of a box has two parameters: the \textbf{side length} and the \textbf{height}. Before the beginning of each episode, a single side length is sampled from a distribution \(\mathbf{Uniform}(\mathbf{SideLength_{low}}, \mathbf{SideLength_{high}})\), and this side length is applied to all the terrain boxes for this episode. The height of each box is sampled from the distribution \(\mathbf{Uniform}(10^{-4} \si{\meter}, \mathbf{H_{max}})\). As all the boxes have the same base side length, they are horizontally densely arranged on the ground without any interval. The maximum box height \(\mathbf{H_{max}}\) determines the difficulty of the terrain. In stage-2 training, \(\mathbf{H_{max}}\) is set to \(\qty{0.12}{\meter}\) constantly for all episodes. In the locomotion performance evaluation, the success rate, average velocity, and other stability indicators are measured at different \(\mathbf{H_{max}}\). Details about the locomotion performance evaluation are in the Experiments section.

\subsection{Robot state initialization}
At the beginning of each episode, we initialize the CPG states, residual angles, robot pose, and user commands as follows. The CPG oscillators are initialized in a random manner. First, a random number \(\theta_a\) is sampled from distribution \(\mathbf{Uniform}(-\pi, \pi)\). Then \(\theta_b = (\theta_a + \pi \,\, \mathbf{if} \,\, \theta_a \leq 0 \,\, \mathbf{else} \,\, \theta_a - \pi) \) is created, i.e., \(\theta_b\) is shifted from \(\theta_a\) by \(\pi\), and \(\theta_b \) is wrapped into \( [-\pi, \pi]\). For all the limbs, initial \(r_i\) is sampled from \(\mathbf{Uniform}(1, 2)\) independently. \(\dot{r}_i\) is initialized to zero. Initial \(\theta_1\) and \(\theta_4\) are set to \(\theta_a\). Initial \(\theta_2\) and \(\theta_3\) are set to \(\theta_b\). Initial \(\phi_i\) is sampled from \(\mathbf{Uniform}(-\frac{\pi}{12}, \frac{\pi}{12})\) independently. The residual angle values are initialized as zeros. The joint angles in MuJoCo state vector are directly set to the target joint angles. 

At initialization, a free-falling initialization is used to create a natural and valid initial pose of the robot in the global world. i.e., the robot is released from a fixed location above the ground, and the robot will do a free-fall until at least one of the feet touches the ground. During the free-fall, the overall target joint positions are fixed to the initial values, and the PD controller runs at each MoJoCo timestep to maintain roughly fixed joint positions. No RL actor inference or training happens during the free-fall. When the contact between the feet and the ground is detected, the velocities of all the DoFs of the robot are set to zero. Therefore, the robot is now frozen so that it has an upright posture with zero velocity and is touching the ground. After one more MuJoCo forward step, the robot is well-prepared to start the first RL control step.

The spatial reference point of the robot is located inside the trunk, which is set up by the robot manufacturer and defined within the MuJoCo model specifications. Before the free-fall, the spatial reference point of the robot is set to \((0,0,0.5) \,\si{\meter}\) in the global frame of the MuJoCo world. In this case, the robot is guaranteed to be above the ground and not to contact the ground, neither on flat ground nor on uneven terrain.

\newpage

\section{EP-based RL training}
\label{sec:EP_A}

The PPO variant adopted in this paper contains three trainable parts: policy network, value network, and the log-std vector. In the work, two EP networks, a type of recurrent neural network with bi-directional connections, instead of standard feed-forward networks, play the role of the CPG policy network and the RES policy network.
The log of action standard deviation is a standalone vector and will be updated individually instead of being part of the policy network output. The EP policy only outputs the mean of action.

\subsection{PPO training process overview}

The entire PPO training process can be split into a sequence of PPO training units. A PPO training unit contains a rollout phase and a neural network update phase. In the rollout phase, the robot is directed to walk/run in each environment for \(T\) control steps. In the case of simulation, the total number of environments is \(N\). At each control step, firstly, an observation vector \(s_t\) is generated by the robot. The observation vector contains external information (for example, user command), and internal information (for example, the IMU sensor and foot contact sensor output). According to the observation vector, the policy network outputs an action mean vector \(\mu_{t}\). In an actual deployment and performance testing situation, this action mean vector is directly used for robot control. However, in the training process, some new actions need to be explored for developing potentially better locomotion strategies. Therefore, in the training process, a noise vector \(\epsilon_{t}\) will be added to the action mean vector, resulting in the action vector \(a_t = \mu_{t} + \epsilon_{t}\). The noise vector \(\epsilon_{t}\) is usually sampled from a zero-mean normal distribution \(\mathcal{N}(0, \sigma)\). Or equivalently, it can be described that the action \(a_t\) is sampled from the distribution \(\mathcal{N}(\mu_{t}, \sigma)\). It should be noted that \(\sigma\) is from a standalone log-std vector. At the same time, the logarithm of the probability of sampling this action vector \(\mathrm{log}({\pi(a_t|s_t)})\) will be recorded. \(\pi(a_t|s_t) = f_{\mathcal{N}(\mu_{t}, \sigma)}(a_t)\), where \(f\) is the probability density function of \(\mathcal{N}(\mu_{t}, \sigma)\). In the rollout dataset, this log probability will be marked as \(\mathrm{log}(\pi_{\mathrm{rollout}}(a_t|s_t))\) because if the policy network is updated, the log probability of \(a_t|s_t\) will change. The ``rollout'' marker distinguishes it from an after-policy-update log probability or current log probability \(\mathrm{log}({\pi(a_t|s_t)})\).  The action vector is applied in the robot control, and the environment time step is forwarded by \(\Delta t_{\mathrm{RL}}\). Then, the reward function outputs a scalar reward \(r_t\) according to the stability, velocity, and energy consumption of the locomotion, and the robot will receive the next step observation vector \(s_{t+1}\). Subsequently, the value function for current and next step observation \(\mathbf{Value}(s_t), \mathbf{Value}(s_{t+1})\) are calculated through value network inference. Finally, from the environment, a done signal \(d_t\), which indicates the end of the current episode, and a robot falling signal \(\mathrm{fall}_t\), which indicates whether the robot fell in the current step, will be collected. It should be noted here that, if the maximum episode length is reached but the robot has not fallen, future reward is assumed non-zero. In this case, the value function of the next state will be used as the bootstrapped future cumulative reward after the final step of the episode. In summary, at time \(t\), the following data tuple will be stored in the buffer
\[
(s_t, a_t, r_t, \mathrm{log}(\pi_{\mathrm{rollout}}(a_t|s_t)), d_t, \mathrm{fall}_t, \mathbf{Value}(s_t), \mathbf{Value}(s_{t+1}))
\]

For the simulation case, if there are \(N\) parallel environments, a total of \(N\cdot T\) data tuples will be collected in the rollout phase.

In the neural network update phase, the data tuples stored in the buffer will be processed. \(\forall t\), first, \(\mathbf{Return}(s_t)
\) and \(\mathbf{Advantage}(s_t, a_t)
\) are calculated through GAE \cite{schulman2015HighDimensionalContinuousControl}, which requires \(r_t, d_t, \mathrm{fall}_t\) and \(\mathbf{Value}(s_t), \mathbf{Value}(s_{t+1})\). 
Afterwards, the policy network and the value network are updated for \(K_\mathrm{epoch}\) epochs and \(\frac{|\mathcal{D}|}{|\mathcal{B}|}\) mini-batches in each epoch, where \(\mathcal{D}\) is a single rollout dataset with \(N\cdot T\) data tuples, and \(\mathcal{B}\) is a data mini-batch.
The policy network is updated by maximizing the PPO action surrogate objective function, which will be discussed in a later section. This step requires \(s_t,\, a_t,\, \mathrm{log}(\pi_{\mathrm{rollout}}(a_t|s_t)), \mathbf{Advantage}(s_t, a_t)\) and \( \sigma_t\). Finally, the value network is updated by minimizing the \(L_2\) loss between the value network output and the return:
\begin{equation}
    \frac{1}{|\mathcal{B}|}\sum_t (\mathbf{Value}(s_t) - \mathbf{Return}(s_t))^2
\label{valueLoss_A}
\end{equation}

\subsection{Backpropagation-based policy update}
The PPO action surrogate objective is
\begin{equation}
L^{\mathrm{CLIP}} = \hat{\mathbb{E}_t} [\min(r_t(\mu_t)\hat{A}_t, \mathbf{Clamp}(r_t(\mu_t), 1-\epsilon, 1+\epsilon)\hat{A}_t)]
\label{policyloss_A}
\end{equation}
Here, \(\hat{A}_t \) is the estimated \( \mathbf{Advantage}(s_t, a_t)\) and \(\epsilon\) is the PPO ratio clip coefficient. \(r_t(\mu_t)\) is the probability ratio between the current action probability \(\pi(a_t|s_t)\) and the action probability stored in the rollout \(\pi_{\mathrm{rollout}}(a_t|s_t)\), i.e.,  \(r_t(\mu_t) = \frac{\pi(a_t|s_t)}{\pi_{\mathrm{rollout}}(a_t|s_t)}\). Then, backpropagation is used to calculate the objective gradient at \(\mu_t\) and further propagate the gradient back and update the policy and value network parameters to maximize the PPO surrogate objective and to minimize the value function loss.

\subsection{EP-based policy network update}

The EP network training contains a free phase and one or more nudge phases \cite{scellier2017equilibrium, laborieux2021scaling}. In different phases, the entire network converges to different equilibrium states determined by the external perturbation from the output layer. In this paper, the one-free-two-nudge version of EP is used for optimal gradient estimation \cite{laborieux2021scaling}. 
For each mini-batch, the policy and value networks individually experience a free phase, a positive nudge phase, and a negative nudge phase. 

For the value network, the loss function is still of the form in Equation \ref{valueLoss_A}, which is a simple regression problem. Only the policy network update will be discussed in detail here.

For the policy network, the input is the observation vector or the pre-processed observation vector. A modified PPO objective gradient is applied to the output layer neurons in the nudge phases. To find the gradient of the PPO action surrogate objective, the probability ratio \(r_t(\mu_t)\) from the objective (Equation \ref{policyloss_A}) can be expanded as:
\begin{equation}
r_t(\mu_t) = \frac{\pi(a_t|s_t)}{\pi_{\mathrm{rollout}}(a_t|s_t)}
= \exp(\mathrm{log}(\pi(a_t|s_t)) - \mathrm{log}(\pi_{\mathrm{rollout}}(a_t|s_t)))
\end{equation}
Given that the dimension \(i\) of action \(a_t\) is sampled from an independent normal distribution with mean \(\mu_{t,i}\) and standard deviation \(\sigma_i\), the probability density of getting action \(a_t\) given observation \(s_t\) is:
\begin{equation}
\pi(a_t|s_t) = \prod_i^{D_{\mathrm{action}}}\left[\frac{1}{\sqrt{2\pi\sigma_i^2}}\mathrm{exp}\left(-\frac{(a_{t,i} - \mu_{t,i})^2}{2\sigma_i^2}\right)\right]
\end{equation}
Here, \(D_{\mathrm{action}}\) is the dimension of the action space. \(\mu_{t,i}\) is the \(i\)-th dimension of the output of the policy network given \(s_t\) and \(\sigma_i\) is the \(i\)-th dimension of the standard deviation vector. It can be derived that the gradient of the probability ratio \(r_t(\mu_t)\) w.r.t. \(\mu_{t,i}\) is:
\begin{align}
\frac{\partial r_t(\mu_t)}{\partial \mu_{t,i}}
&= \frac{\pi(a_t|s_t)}{\pi_{\mathrm{rollout}}(a_t|s_t)} \frac{\partial \mathrm{log}\pi(a_t|s_t)}{\partial \mu_{t,i}} \\
&= \frac{\pi(a_t|s_t)}{\pi_{\mathrm{rollout}}(a_t|s_t)} \frac{(a_{t,i} - \mu_{t,i})}{\sigma_i^2} \notag 
\end{align}
Note that the probability ratio \(\frac{\pi(a_t|s_t)}{\pi_{\mathrm{rollout}}(a_t|s_t)}\) is always close to 1 in the training process. Thus, we can write:
\begin{align}
\frac{\partial r_t(\mu_t)}{\partial \mu_{t,i}} \approx \frac{(a_{t,i} - \mu_{t,i})}{\sigma_i^2}
\end{align}
Therefore, if \(\mu_{t,i}\) is the \(i\)-th dimension of the action mean in sample \(t\), we have the approximate PPO objective gradient w.r.t. the policy network output \(\mu_{t,i}\):
\begin{align}
\label{clip_obj_grad_A}
\frac{\partial L^{\mathrm{CLIP}}}{\partial \mu_{t,i}} &= \frac{1}{|\mathcal{B}|}
    \begin{cases}
      \mathbf{1}_{[0, 1+\epsilon
      )}(r_t(\mu_{t}))\cdot r_t(\mu_{t})\frac{(a_{t,i} - \mu_{t,i})}{\sigma_i^2}\hat{A}_t & \text{if \(\hat{A}_t \geq 0\)}\\
      \mathbf{1}_{(1-\epsilon,
      \infty)}(r_t(\mu_{t}))\cdot r_t(\mu_{t})\frac{(a_{t,i} - \mu_{t,i})}{\sigma_i^2}\hat{A}_t & \text{if \(\hat{A}_t < 0\)}
    \end{cases} \\
    &\approx \frac{1}{|\mathcal{B}|}
    \begin{cases}
      \mathbf{1}_{[0, 1+\epsilon
      )}(r_t(\mu_{t}))\cdot \frac{(a_{t,i} - \mu_{t,i})}{\sigma_i^2}\hat{A}_t & \text{if \(\hat{A}_t \geq 0\)}\\
      \mathbf{1}_{(1-\epsilon,
      \infty)}(r_t(\mu_{t}))\cdot \frac{(a_{t,i} - \mu_{t,i})}{\sigma_i^2}\hat{A}_t & \text{if \(\hat{A}_t < 0\)}
    \end{cases} \notag 
\end{align}
Here, the terms \(\mathbf{1}_{[0, 1+\epsilon)}(r_t(\mu_{t}))\) and \(\mathbf{1}_{(1-\epsilon,\infty)}(r_t(\mu_{t}))\) can be regarded as gradient masks. It sets the continuous part of the gradient \(\frac{(a_{t,i} - \mu_{t,i})}{\sigma_i^2}\hat{A}_t\) to zero when the probability ratio is beyond the gradient enabling range (positive advantage: \([0, 1+\epsilon)\), negative advantage: \((1-\epsilon,\infty)\)).

By using an EP network to optimize the PPO objective, we can apply the PPO objective gradient w.r.t. the network output \(\frac{\partial L^{\mathrm{CLIP}}}{\partial \mu_{t,i}}\) to the output layer neurons in the EP network. The \(\mu_{t,i}\) is the policy network output dimension \(i\) in BP-trained RL, which is equivalent to \(\xi^*_{t,out,i}\) in EP. \(\xi^*_{t,out,i}\) is the output neuron state \(i\) in free phase equilibrium when the policy input observation is \(s_t\).
\textbf{It should be noted here that the subscripts \(t\) in \(\xi^*_{t,out,i}\) represent neither EP relaxation step index nor the time in EP neuron state dynamics, instead, it represents the rollout data sample index, like the \(t\) in \(\mu_{t}\).}
The output neuron state dynamics modulated by the PPO objective gradient is (see main text):
\begin{align}
    \label{EP_out_dynamics_unmodified_A}
    \frac{d\xi_{t,out,i}}{d[\mathrm{time}]} = &
    -\xi_{t,out,i} 
    + \rho'(\xi_{t,out,i})\sum_{j}w_{out,ij} \rho(\xi_{t,j}) + b_{out,i} \rho'(\xi_{t,out,i})\\
    &- \beta\cdot \frac{1}{|\mathcal{B}|}
    \begin{cases}
      \mathbf{1}_{[0, 1+\epsilon
      )}(r_{t,\mathrm{nudging}}(\xi_{t,out}))\cdot \frac{(a_{t,i} - \xi_{t,out,i})}{\sigma_i^2}\hat{A}_t & \text{if \(\hat{A}_t \geq 0\)}\\
      \mathbf{1}_{(1-\epsilon,
      \infty)}(r_{t,\mathrm{nudging}}(\xi_{t,out}))\cdot \frac{(a_{t,i} - \xi_{t,out,i})}{\sigma_i^2}\hat{A}_t & \text{if \(\hat{A}_t < 0\)}
    \end{cases} \notag 
\end{align}
However, based on experiments, this EP output neuron dynamics (Equation \ref{EP_out_dynamics_unmodified_A}) leads to failure in training convergence. The per-update KL-divergence graph indicates excessively large update steps. We investigated the cause of the large update steps. In conventional PPO, backpropagation directly calculates the parameter gradient. But the nudge phases of EP move the output layer neurons in two-directions: the \textbf{closer-to-target} direction or the \textbf{farther-away-from-target} direction. For example, according to Equation \ref{EP_out_dynamics_unmodified_A}, when \(\beta = \beta_{\mathrm{ep}}\) and \(\hat{A}_t > 0\), the output neurons are moving in a direction that increase its distance from the action sample, because the objective gradient is proportional to \(\xi_{t,out,i} - a_{t,i}\). More concretely, before the very first positive nudge step
\begin{equation}
    -\xi_{t,out,i} 
    + \rho'(\xi_{t,out,i})\sum_{j}w_{out,ij} \rho(\xi_{t,j}) + b_{out,i} \rho'(\xi_{t,out,i}) = 0
\end{equation}
Therefore,
\begin{equation}
    \frac{d\xi_{t,out,i}}{d[\mathrm{time}]} \propto (\xi_{t,out,i} - a_{t,i}) \quad \text{before first positive nudge phase step}
\end{equation}
If the output neuron state shift along the farther-away-from-target direction is not bounded, \(r_{t,{\mathrm{nudging}}}(\xi_{t,out})\) can be indefinitely close to zero. 
Similarly, when \(\beta = \beta_{\mathrm{ep}}\) and \(\hat{A}_t < 0\), there is no upper bound for \(r_{t,{\mathrm{nudging}}}(\xi_{t,out})\), but the normal distribution has a probability density upper bound, which does not allow the value explosion of \(r_{t,{\mathrm{nudging}}}(\xi_{t,out})\).
When \(\beta = -\beta_{\mathrm{ep}}\) and \(\hat{A}_t < 0\), the driving force is also farther-away-from-target. However, ratio lower bound exists in this case: \(\mathbf{1}_{(1-\epsilon,\infty)}(r_{t,{\mathrm{nudging}}}(\xi_{t,out}))\). 
In backpropagation, there is not an equivalent concept of this reverse driving force. Therefore this is an \textbf{EP-only phenomenon}.

To constrain these excessively large shifts, we propose the two-sided PPO ratio clip objective gradient:
\begin{equation}
\label{modified_policy_loss_grad_A}
\frac{\partial L^{\mathrm{TwoSidedCLIP}}}{\partial \xi_{t,out,i}} = \frac{1}{|\mathcal{B}|}
    \begin{cases}
      \mathbf{1}_{(1-\epsilon_{\mathrm{rev}}, 1+\epsilon
      )}(r_{t,\mathrm{nudging}}(\xi_{t,out}))\cdot \frac{(a_{t,i} - \xi_{t,out,i})}{\sigma_i}\hat{A}_t & \text{if \(\hat{A}_t \geq 0\)}\\
      \mathbf{1}_{(1-\epsilon,
      1+\epsilon_{\mathrm{rev}})}(r_{t,\mathrm{nudging}}(\xi_{t,out}))\cdot \frac{(a_{t,i} - \xi_{t,out,i})}{\sigma_i}\hat{A}_t & \text{if \(\hat{A}_t < 0\)}
    \end{cases} 
\end{equation}
Here, \(\epsilon_{\mathrm{rev}}\) is the reverse PPO ratio clip coefficient. The non-zero gradient is applied to the output layer only when the output neuron state is within the range of the two-sided bounded gradient enabling range. 

Additionally, the exploration noise standard deviation \(\sigma_i\), in the conventional \([-1,1]\) action space, is usually smaller than \(1\) and can potentially result in a very large magnitude of the gradient due to the inverse quadratic variation, since the equilibrium shift associated with EP learning is very sensitive to the gradient magnitude.

Our experiments show that this two-sided bounded gradient and reduced gradient scaling factor \(\frac{1}{\sigma_i}\) successfully leads to training convergence. The ablation studies involving different values of \(\epsilon_{\mathrm{rev}}\) and \(\frac{1}{\sigma_i}\) are shown in Appendix~\ref{sec:ablation_study}.

\subsection{Log-std vector update}
To prevent using backpropagation in the entire algorithm, the gradient of the PPO objective and entropy loss w.r.t. the log-std vector is also explicitly calculated. It needs to be noted that the log-std vector here has the same dimensionality as the action space, which is 12. Note that the original PPO objective is used. The PPO objective gradient w.r.t. the \(i\)-th dimension of the log-std vector is
\begin{equation}
\label{obj_grad_wrt_logstd_A}
\frac{\partial L^{\mathrm{CLIP}}}{\partial \mathrm{log}(\sigma_{i})} = \frac{1}{|\mathcal{B}|} \sum_t
    \begin{cases}
      \mathbf{1}_{[0, 1+\epsilon
      )}(r_t(\sigma))
      \cdot \left[\frac{(a_{t,i} - \mu_{t,i})^2}{\sigma_i^2}-1\right]
      r_t(\sigma)\hat{A}_t & \text{if \(\hat{A}_t \geq 0\)}\\
      \mathbf{1}_{(1-\epsilon,
      \infty)}(r_t(\sigma))
      \cdot \left[\frac{(a_{t,i} - \mu_{t,i})^2}{\sigma_i^2}-1\right]
      r_t(\sigma)\hat{A}_t & \text{if \(\hat{A}_t < 0\)}
    \end{cases} 
\end{equation}
Here, the probability ratio \(r_t\) is regarded as a function of \(\sigma\), as we only consider the effect of the shift of \(\sigma\) on the shift of \(r_t\).
\begin{equation}
r_t(\sigma) = \frac{\pi(a_t|s_t)}{\pi_{\mathrm{rollout}}(a_t|s_t)}
= \mathrm{exp}(\mathrm{log}(\pi(a_t|s_t)) - \mathrm{log}(\pi_{\mathrm{rollout}}(a_t|s_t)))
\end{equation}
The entropy loss w.r.t. the \(i\)-th dimension of the log-std vector is
\begin{equation}
\label{entropy_grad_wrt_logstd_A}
\frac{\partial L_{\mathrm{entropy}}}{\partial \mathrm{log}(\sigma_{i})} = 
2\cdot k_{\mathrm{entropy}}\cdot (H(\mathrm{log}(\sigma)) - H_{\mathrm{target}})
\end{equation}
Here, \(k_{\mathrm{entropy}}\) is the entropy loss coefficient, and \(H_{\mathrm{target}}\) is the target entropy determined by the entropy scheduler.
The entropy of a normal distribution with an uncorrelated per-dimension-log-std vector is:
\begin{equation}
\label{entropy_function_A}
H(\mathrm{log}(\sigma)) = \frac{1}{2} \sum_{i=1}^{D_{\mathrm{action}}}(\mathrm{log}(2\pi e) + 2 \mathrm{log}(\sigma_{i}))
\end{equation}
The total gradient applied to the \(i\)-th dimension of log-std vector is
\begin{equation}
\label{logstd_grad_A}
    \frac{\partial L^{\mathrm{CLIP}}}{\partial \mathrm{log}(\sigma_{i})} + \frac{\partial L_{\mathrm{entropy}}}{\partial \mathrm{log}(\sigma_{i})}
\end{equation}
In fact, the entropy loss gradient w.r.t. each dimension of the log-std vector is identical.

\subsection{EP network architecture}
Both policy and value networks receive a 1024-D iDCT converted observation vector. Both networks have 2 hidden layers and 768 neurons in each hidden layer. The policy network outputs a 12-D action mean, and the value network outputs a 1-D value function. The EP networks use bias for each neuron in the hidden and output layers. Neuron activation function applied to the hidden layer neurons is hard-sigmoid (\(\mathbf{Clamp(\cdot, 0, 1)}\)). No activation function is applied to the input and output layer neurons. The weight matrix in the EP network is initialized through sampling each weight element independently from the following distribution:
\begin{equation}
    \mathbf{Uniform}\left(-\frac{\alpha_{w}}{\sqrt{[\#\mathrm{Fan\_left}]}},\frac{\alpha_{w}}{\sqrt{[\#\mathrm{Fan\_left}]}}\right)
\end{equation} 
Here, \(\alpha_w\) is the EP weight initialization scale, and \([\#\mathrm{Fan\_left}]\) represents the number of neurons in left layer. From the perspective of a weight matrix, the left layer represents the neuron layer that is closer to the input layer.

\newpage

\section{Details about walking test and extended experiments}
\label{sec:extended_experiments}

\subsection{Walking test setup}
In the 5-meter walking test proposed in \cite{seto2025two}, the robot is expected to walk for 5 meters within 30 seconds with \(v_{x,\mathrm{local,target}} = \qty{0.3}{\meter/\second}\). The distance is calculated by integrating \(v_{x,\mathrm{local}}\) across time. If the robot successfully reaches 5 meters within 30 seconds without falling, this trial is marked as a success. In cases where the robot falls or cannot reach 5 meters within 30 seconds, this trial is marked as a failure. The test is repeated under the condition of 7 different \(\mathbf{H_{max}}\) in the set \(\{0, 0.02, 0.04, 0.06, 0.08, 0.10, 0.12\}\si{\meter}\).

In this paper, we used an extended version of the 5-meter walking test by testing the locomotion policy with three different local x-directional target velocity \(v_{x,\mathrm{local,target}} \in \{0.1, 0.3, 0.5\}\si{\meter/\second}\). For different local x-directional target velocities, the target traveling distance is also proportionally scaled, i.e., when \(v_{x,\mathrm{local,target}} = \qty{0.1}{\meter/\second}\), the target traveling distance is 
\(\qty{5}{\meter} \times (\qty{0.1}{\meter/\second})/(\qty{0.3}{\meter/\second}) = \qty{1.667}{\meter}\). Similarly, the target traveling distance for \(v_{x,\mathrm{local,target}} = \qty{0.5}{\meter/\second}\) is \(\qty{8.333}{\meter}\). We also removed \(\qty{0}{\meter}\) from the list of \(\mathbf{H_{max}}\) as the flat ground performance will always be better than any non-zero \(\mathbf{H_{max}}\). All the testing combinations is the following product set:
\begin{equation}
    \{\mathrm{EP},\mathrm{BP}\} \times\{0.1, 0.3, 0.5\}\si{\meter/\second} \times \{0.02, 0.04, 0.06, 0.08, 0.10, 0.12\}\si{\meter}
\end{equation}
The following locomotion performance indicators are recorded: success rate, average forward velocity, average actuator total power, average absolute roll/pitch angle, and average absolute roll/pitch angular velocity. 

Each of the 5 versions is tested individually.
For each version, under the condition of each combination, we run 500 episodes of uneven terrain locomotion. It should be noted that the length of each episode could be different. For each version and each combination of (EP/BP, \(v_{x,\mathrm{local,target}}\), \(\mathbf{H_{max}}\)), we calculated the success rate for the 500 episodes. We also calculated the weighted average of other performance indicators. The weight is the time duration of the episode. For example, a 20-second-long episode has a proportionally larger weight than a 5-second-long episode.

\begin{figure}[H]
    \centering
    \includegraphics[width=1.0\textwidth]{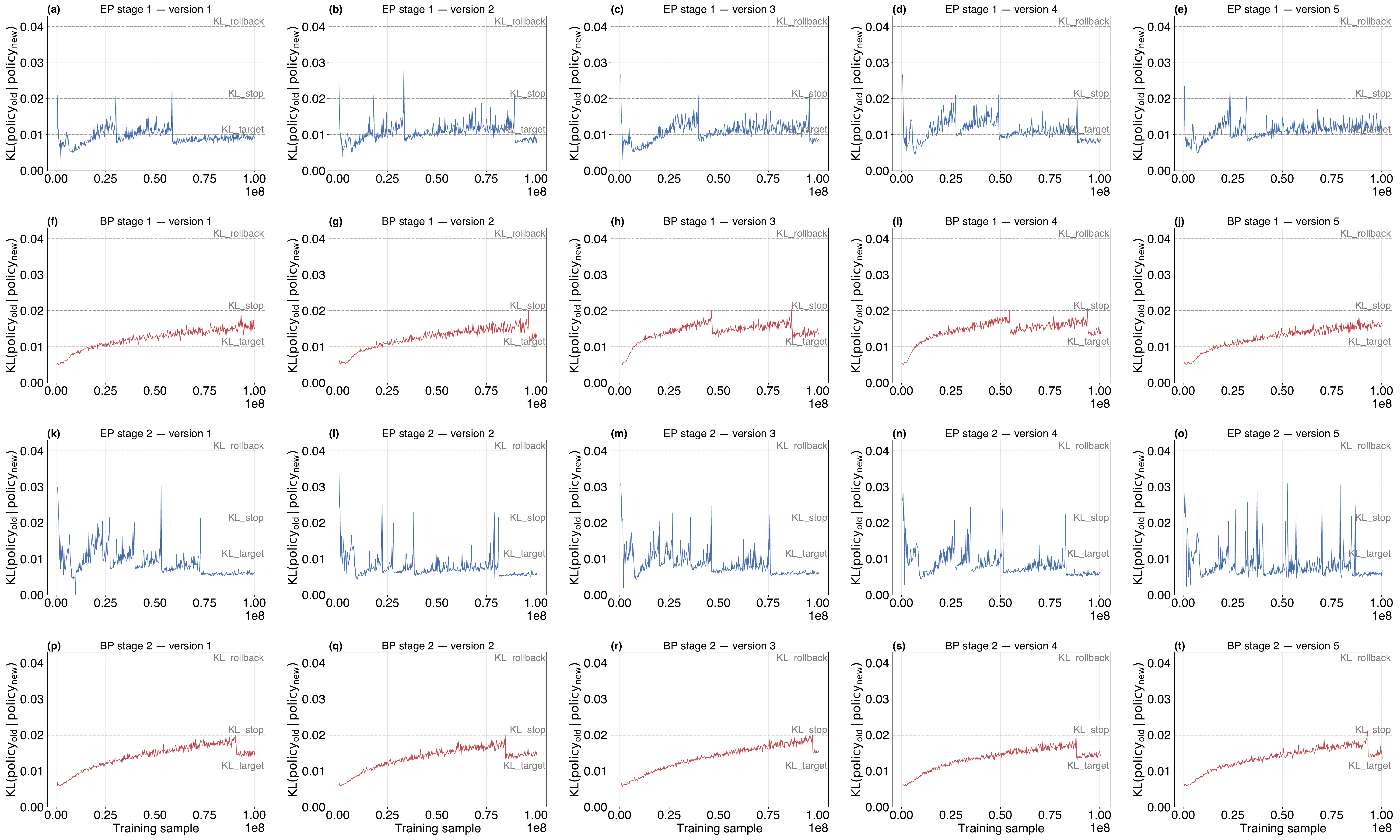}
    \caption{Per-update KL-divergence. The KL-divergence of all 5 versions, EP and BP algorithms, and both stages are plotted here. Here, the y-axis label \(\text{KL}({\text{policy}_\text{old}}|{\text{policy}_\text{new}})\) is the KL-divergence between the pre-update and post-update policy.}
    \label{kl_div_all_version}
\end{figure}

\subsection{Per-update KL-divergence in EP and BP-based training process}

The main paper shows the figure for the rollout mean reward and value MSE in the training process of EP and BP-based locomotion training. In Fig. \ref{kl_div_all_version}, the per-update KL-divergence (KL-div) for all 5 versions of stage-1 and stage-2 training is plotted. It is clear that BP yields a smoother increment of KL-div in the training process, while EP-based training hits the early stop KL threshold multiple times. The hypothetical cause is that the EP policy network has a non-linear response to the objective gradient, and a small change in the objective gradient leads to a large difference in the KL-div.

\subsection{EP nudge phase dynamics with PPO objective gradient}
Given an observation vector \(s_t\), the policy network returns an action mean vector \(\mu_t\). The intuitive idea of PPO is that, if, for observation \(s_t\), there is any action \(a_t\) satisfying \(\mathbf{Advantage}(s_t, a_t) > 0\), make \(\mu_t\) be closer to \(a_t\). Conversely, if \(\mathbf{Advantage}(s_t, a_t) < 0\), make \(\mu_t\) be farther away from \(a_t\). Also, it is critical to maintain small update steps.

This leads us to the question: if we update the policy network according to \(\mathbf{Advantage}(s_t, a_t)\) for a group of samples, by how much will corresponding \(\mu_t\) shift? The intuition is that \(\mu_t\) will shift significantly if \(\mathbf{Advantage}(s_t, a_t)\) has larger magnitude, and \(\mu_t\) will have minimal shift if \(\mathbf{Advantage}(s_t, a_t)\) is close to zero. From the perspective of the PPO, the change of \(\frac{\pi(a_t|s_t)}{\pi_{\mathrm{rollout}}(a_t|s_t)}\) should be considered instead of the shift of \(\mu_t\). \(\pi_{\mathrm{rollout}}(a_t|s_t)\) gives a reference point during policy update: the updated policy should not be very distant from the reference point. The advantage-probability ratio relationship is the core of PPO. Excessively large update step yields policy collapse. For EP-based PPO, investigating the advantage-\(r_{\mathrm{nudging}}\) relationship is important for maintaining stable policy improvement. 
\begin{align}
\label{r_nudging_A}
    r_{t,\mathrm{nudging}}(\xi_{t,out}) &= \frac{\pi_{\mathrm{nudging}}(a_t|s_t)}{\pi_{\mathrm{rollout}}(a_t|s_t)}\\
    &=\frac{1}{\pi_{\mathrm{rollout}}(a_t|s_t)} \prod_i^{D_{\mathrm{action}}}\left[\frac{1}{\sqrt{2\pi\sigma_i^2}}\mathrm{exp}\left(-\frac{(a_{t,i} - \xi_{t,out,i})^2}{2\sigma_i^2}\right)\right] \notag
\end{align}
According to the definition of \(r_{t,\mathrm{nudging}}(\xi_{t,out})\) in Equation \ref{r_nudging_A}, \(r_{t,\mathrm{nudging}}(\xi_{t,out})\) will change across relaxation steps in EP network nudge phases. In the positive nudge phase, if the advantage is greater than zero, \(r_{t,\mathrm{nudging}}(\xi_{t,out})\) should decrease to a value that is less than 1, i.e., the \(\pi_{\mathrm{nudging}}(a_t|s_t)\) is expected to be smaller than the \(\pi_{\mathrm{rollout}}(a_t|s_t)\).

To further analyze the relationship between the normalized advantage and \(r_{t,\mathrm{nudging}}(\xi_{t,out})\), we recorded the policy output-layer state \(\xi_{out}\) during the nudge phases. The \(r_{t,\mathrm{nudging}}(\xi_{t,out})\) is calculated by using the action, rollout probability, and log-std in the data analysis. In a training process, after a specific rollout data collection, two separate testing nudge phases were conducted. We randomly chose 1000 samples from the rollout dataset. In these two nudge phases, the value of \(\xi_{t, out}\) for each sample \(t\) was recorded for each relaxation step. Through this pair of nudge phases, we can know how \(\xi_{t,out}\) is changing in the nudge phases according to its advantages. As these 1000 samples are randomly selected, their normalized advantage is approximately normally distributed. 

\begin{figure}[h]
    \centering
    \includegraphics[width=1.0\textwidth]{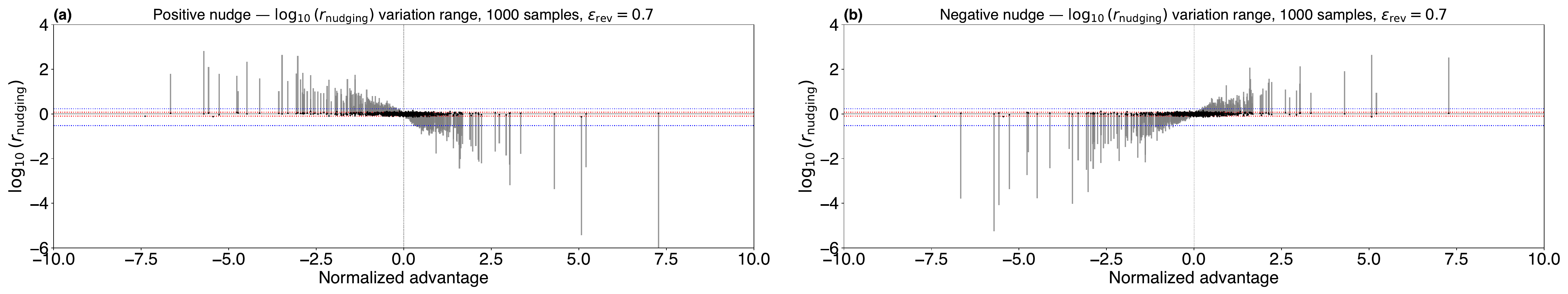}
    \caption{The advantage-\(r_{\mathrm{nudging}}\) relationship of 1000 samples in the middle of a training process. The variation of \(\log_{10}(r_{t,\mathrm{nudging}}(\xi_{t,out}))\) in positive and negative nudge phases are shown. The black dots represent the \(\log_{10}(r_{t}(\xi^*_{t,out}))\) at free phase equilibrium. The gray line segments represent variation range of \(\log_{10}(r_{t,\mathrm{nudging}}(\xi_{t,out}))\) in the entire positive or negative phase.}
    \label{nudge_log10_ratio_A}
\end{figure}

\begin{figure}[h!]
    \centering
    \includegraphics[width=1.0\textwidth]{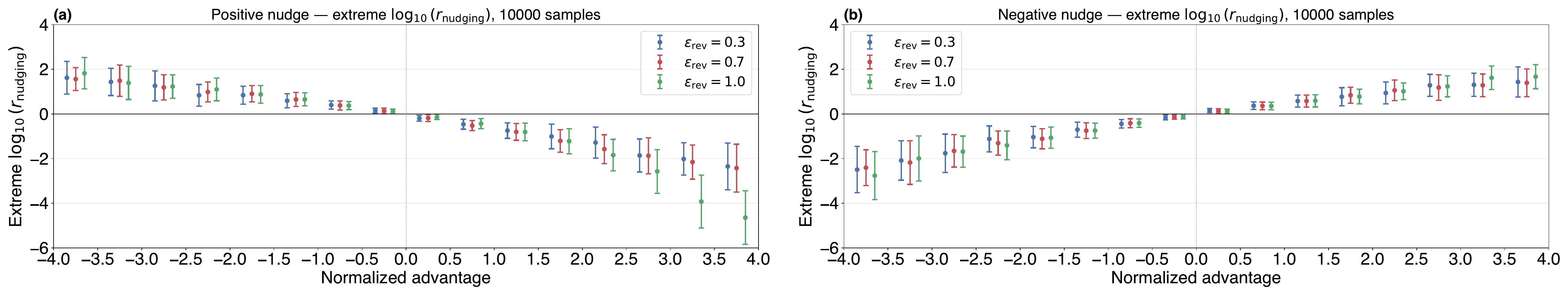}
    \caption{Statistics of the aggregated extreme \(\log_{10}(r_{t,\mathrm{nudging}}(\xi_{t,out}))\) across 10 rollouts. It can be observed that in the left panel, in the normalized advantage range \([2.5,4]\), the extreme \(\log_{10}(r_{t,\mathrm{nudging}}(\xi_{t,out}))\) is significantly lower for \(\epsilon_{\mathrm{rev}}=1.0\) compared with two other cases.}
    \label{nudge_extreme_statistics_A}
\end{figure}

Thus, we can pose the following questions: for a sample with specific advantage \(\hat{A}_t\), by how much will the corresponding \(\xi_{t,out}\) change in the nudge phase? In which direction will \(\xi_{t,out}\) change? By how much will \(r_{t,\mathrm{nudging}}(\xi_{t,out})\) change with \(\xi_{t,out}\)? Our experiments show that, although the after-update probability ratio \(r_t(\xi_{t,out})\) changes very slightly, in the nudge phases, \(r_{t,\mathrm{nudging}}(\xi_{t,out})\) shifts significantly. \(r_{t,\mathrm{nudging}}(\xi_{t,out})\) can reach values as low as \(10^{-5}\) or as high as \(10^{3}\), although the PPO ratio clip coefficient is 0.2 (which tries to limits the \(r_t(\xi_{t,out})\) to a range \([0.8, 1.2]\)). We observed that larger the magnitude of the advantage, larger is the change of \(r_{t,\mathrm{nudging}}(\xi_{t,out})\). However, \(r_{t,\mathrm{nudging}}(\xi_{t,out})\) is always positive because it is a ratio between probability densities. It is easy to visualize the change of \(r_{t,\mathrm{nudging}}(\xi_{t,out})\) in a log scale. In Fig. \ref{nudge_log10_ratio_A}, \(\log_{10}(r_{t,\mathrm{nudging}}(\xi_{t,out}))\) is plotted versus the advantage \(\hat{A}_t\). It can be observed that \(\log_{10}(r_{t,\mathrm{nudging}}(\xi_{t,out}))\) changes approximately proportionally to the advantage with different coefficients in positive and negative phases. Actually, \(\xi_{t,out}\) and \(r_{t,\mathrm{nudging}}(\xi_{t,out})\) keep oscillating in the nudging phase. This phenomenon is consistent with the nudge driving force in Equation \ref{modified_policy_loss_grad_A}.
To understand the characteristics of this advantage-\(r_{\mathrm{nudging}}\) relationship more clearly, we collected more nudge phase data. Instead of a single rollout, we selected 10 consecutive rollouts. Similarly, in each rollout update, 1000 randomly selected samples of data are used for two nudge phases. Therefore, in total, 10000 samples are collected for the statistics of the advantage-\(r_{\mathrm{nudging}}\) relationship. 
The effect of three different \(\epsilon_{\mathrm{rev}}\) are also investigated. The data for each \(\epsilon_{\mathrm{rev}}\) are aggregated into 16 advantage bins spanning \([-4,4]\). Within each bin, for each \(\epsilon_{\mathrm{rev}}\), the mean and standard deviation of extreme \(\log_{10}(r_{t,\mathrm{nudging}}(\xi_{t,out}))\) are calculated. Extreme \(\log_{10}(r_{t,\mathrm{nudging}}(\xi_{t,out}))\) means the ever reached most distant \(\log_{10}(r_{t,\mathrm{nudging}}(\xi_{t,out}))\) value from zero. Intuitively, extreme \(\log_{10}(r_{t,\mathrm{nudging}}(\xi_{t,out}))\) represent the tip of the gray line segments in Fig. \ref{nudge_log10_ratio_A}. 
Fig. \ref{nudge_extreme_statistics_A} shows the statistics of the extremes of \(\log_{10}(r_{t,\mathrm{nudging}}(\xi_{t,out}))\) within each aggregation bin for 10000 samples in total.

From Fig. \ref{nudge_extreme_statistics_A}, we can see that the advantage-\(r_{\mathrm{nudging}}\) relationship is approximately the same for different \(\epsilon_{\mathrm{rev}}\). However, a critical difference is present in the region \(\hat{A}_t \in [2.5, 4.0]\) in positive nudge phases: the \(\epsilon_{\mathrm{rev}}=1.0\) setting gives significantly lower extreme \(\log_{10}(r_{t,\mathrm{nudging}}(\xi_{t,out}))\) than \(\epsilon_{\mathrm{rev}}=0.3\) and \(\epsilon_{\mathrm{rev}}=0.7\).

To understand this phenomenon, consider the following equation in the positive nudge phase, and when the advantage is positive:
\begin{align}
    \frac{\partial\xi_{t,out,i}}{\partial t} = &
    -\xi_{t,out,i} 
    + \rho'(\xi_{t,out,i})\sum_{j}w_{out,ij} \rho(\xi_{t,j}) + b_{out,i} \rho'(\xi_{t,out,i})\\
    &- \beta\cdot \mathbf{1}_{(1-\epsilon_{\mathrm{rev}}, 1+\epsilon)}(r_{t,\mathrm{nudging}}(\xi_{t,out}))\cdot 
      \frac{(a_{t,i} - \xi_{t,out,i})}{\sigma_i}\hat{A}_t \notag
\end{align}
The nudge force is driving \(\xi_{out}\) away from the sample action. However, given \(r_{t,\mathrm{nudging}}(\xi_{t,out}) > 0\) and \(\epsilon_{\mathrm{rev}}=1.0 \Leftrightarrow 1-\epsilon_{\mathrm{rev}} = 0\)
\[
\mathbf{1}_{[0, 1+\epsilon)}(r_{t,\mathrm{nudging}}(\xi_{t,out}))
\]
cannot bound \(r_{t,\mathrm{nudging}} (\xi_{t,out})\) shift in the negative direction. \(r_{t,\mathrm{nudging}}(\xi_{t,out})\) can be indefinitely close to zero. For all other \(\epsilon_{\mathrm{rev}} < 1.0\), \(r_{t,\mathrm{nudging}}(\xi_{t,out})\) shift is bounded in the negative direction. The discussion about other scenarios: (positive nudge/negative advantage), (negative nudge/positive advantage), (negative nudge/negative advantage) is in Appendix~\ref{sec:EP_A}.

\newpage

\section{Ablation studies}
\label{sec:ablation_study}

\subsection{Effect of different reverse PPO ratio clip coefficients}

\begin{figure}[H]
    \centering
    \includegraphics[width=1.0\textwidth]{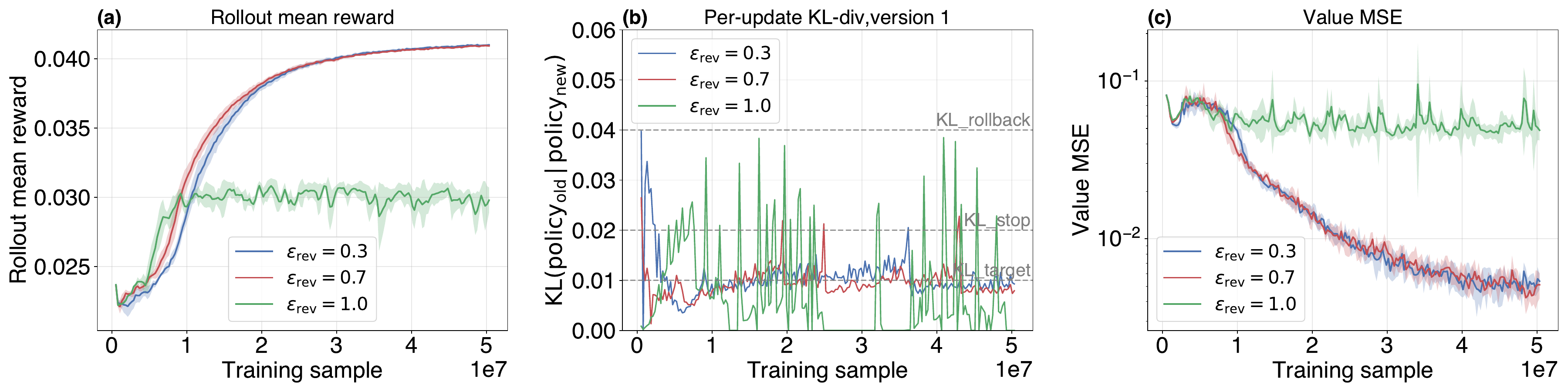}
    \caption{Convergence of stage-1 training for three values of \(\epsilon_{\mathrm{rev}}\). The center line and shaded region indicate the mean and standard deviation across 5 versions for \(\epsilon_{\mathrm{rev}}=0.3\) and \(\epsilon_{\mathrm{rev}}=0.7\). For \(\epsilon_{\mathrm{rev}}=1.0\), only 4 versions are aggregated because there is one version resulting in a numerical instability issue during training.}
    \label{different_rev_convergence_plot_A}
\end{figure}

Fig. \ref{different_rev_convergence_plot_A} shows the rollout mean reward, KL-div, and the value MSE in the training process for three values of \(\epsilon_{\mathrm{rev}}\). It can be observed that \(\epsilon_{\mathrm{rev}} = 0.3 \) and \(\epsilon_{\mathrm{rev}}=0.7\) achieve similar convergence speed, and they both converge to the optimum. \(\epsilon_{\mathrm{rev}}=1.0\) failed to converge. Also, one version of \(\epsilon_{\mathrm{rev}}=1.0\) results in a numerical stability issue in the training process, i.e., the EP network parameters all become NaN. Therefore, we only collect and plot the mean and standard deviation for 4 versions of \(\epsilon_{\mathrm{rev}}=1.0\). From Fig. \ref{different_rev_convergence_plot_A} (b), we can see spiking of per-update KL-div and frequent update rollback, which is indicated by zero value KL-div, which is a sign of excessively large update step. The cause of failed convergence is comprehensively discussed in the Methods section and Appendix~\ref{sec:EP_A}.

\subsection{Effect of \(\frac{1}{\sigma}\) and \(\frac{1}{\sigma^2}\) in the objective gradient}

\begin{figure}[H]
    \centering
    \includegraphics[width=1.0\textwidth]{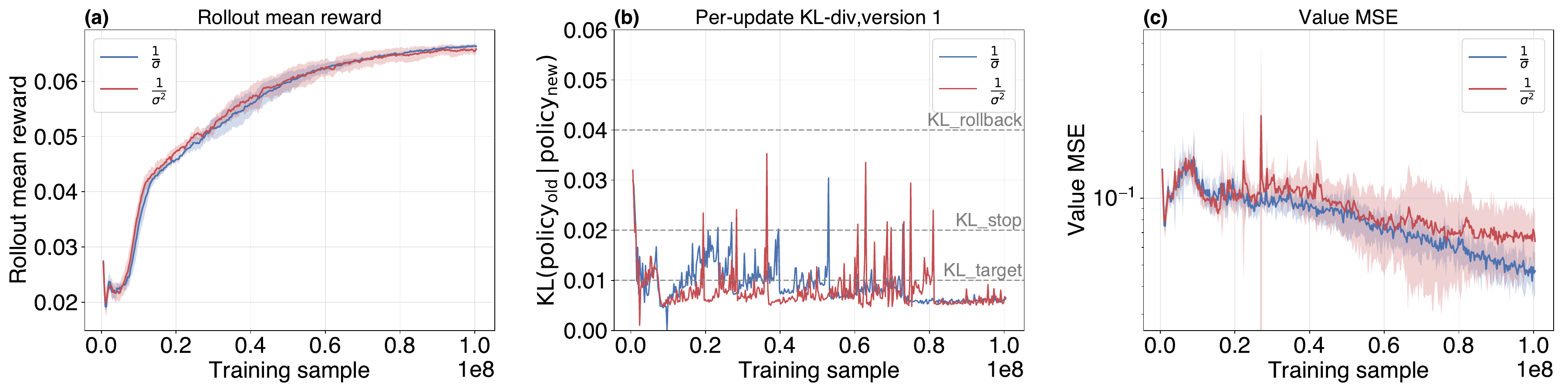}
    \caption{Convergence of stage-2 training for gradient scaling factor \(\frac{1}{\sigma}\) and \(\frac{1}{\sigma^2}\). The center line and shaded region indicate the mean and standard deviation across 5 versions.} 
    \label{sigma_1_vs_sigma_2_A}
\end{figure}

Fig. \ref{sigma_1_vs_sigma_2_A} shows the rollout mean reward, KL-div, and the value MSE in the training process when the output-gradient scaling uses \(\frac{1}{\sigma}\) or \(\frac{1}{\sigma^2}\). While \(\frac{1}{\sigma}\) and \(\frac{1}{\sigma^2}\) yield similar speed of mean reward improvement, and both of them approach the optimal policy, \(\frac{1}{\sigma^2}\) leads to more frequent spikes in the KL-div graph, and the value MSE is not being minimized stably. This phenomenon implies an unstable policy network update. The cause is discussed in Appendix~\ref{sec:EP_A}. Therefore, we choose \(\frac{1}{\sigma}\) for the proposed new objective gradient.

\subsection{Effect of dynamic and static gradient mask}

\begin{figure}[H]
    \centering
    \includegraphics[width=1.0\textwidth]{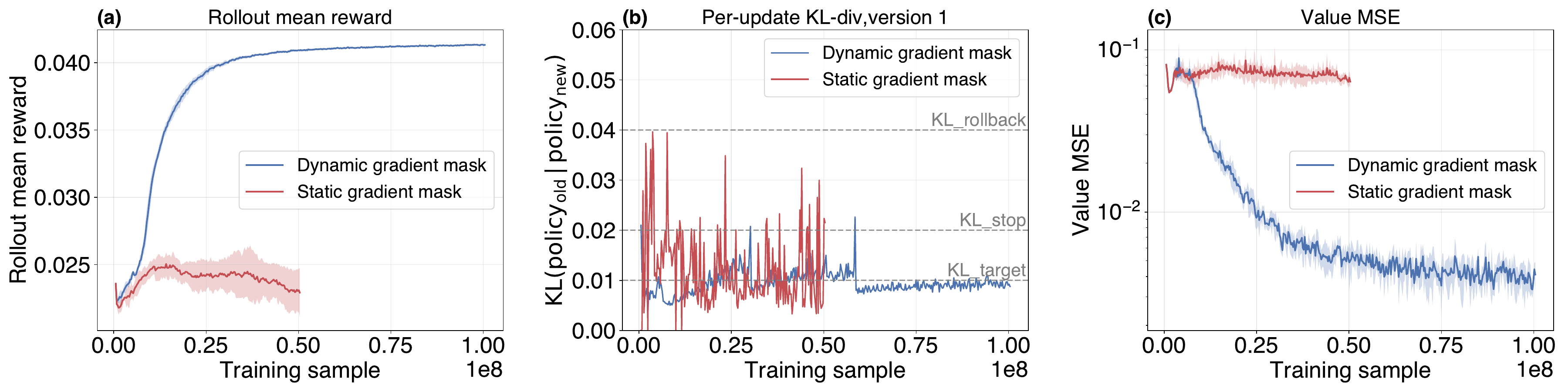}
    \caption{Convergence of stage-1 training for dynamic and static gradient mask. The center line and shaded region indicate the mean and standard deviation across 5 versions.}
    \label{dynamic_static_gradient_mask_A}
\end{figure}

Fig. \ref{dynamic_static_gradient_mask_A} shows the rollout mean reward, KL-div, and the value MSE in the training process for the dynamic and static gradient mask. The definition of static gradient masks is as follows. When using a static gradient mask, the objective gradient applied to the output neuron state \(\xi_{t,out,i}\) is:  
\begin{equation}
    \mathbf{StaticMask}_t\frac{1}{|\mathcal{B}|} \frac{(a_{t,i} - \xi_{t,out,i})}{\sigma_i^2}\hat{A}_t
\end{equation}
A constant gradient mask \(\mathbf{StaticMask}_t = \mathbf{1}_{[0, 1+\epsilon)}(r_{t}(\xi_{t,out}))\) or \(\mathbf{StaticMask}_t = \mathbf{1}_{(1-\epsilon, \infty)}(r_{t}(\xi_{t,out}))\) is calculated based on the free phase equilibrium output neuron state before the nudge phase. 

This constant mask is used in the nudge phase without changing. The static gradient masks mimic the backpropagation-style gradient masks, which are calculated only once. 
It can be clearly noticed that a static mask-based objective gradient cannot improve the policy. For a static gradient mask-based objective gradient, only \(5\times 10^7\) samples are trained.

\subsection{Effect of iDCT observation preprocessing}
\begin{figure}[H]
    \centering
    \includegraphics[width=1.0\textwidth]{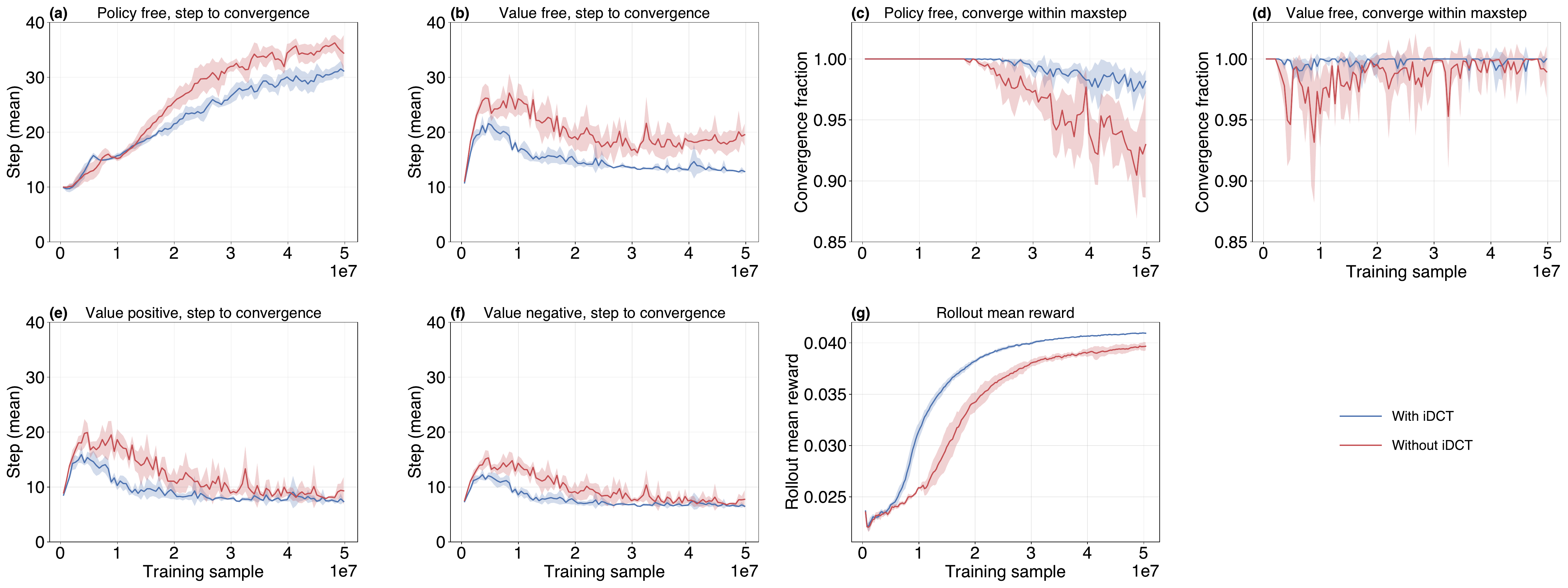}
    \caption{Step to convergence in the EP relaxation for both policy and value networks. The test is conducted in stage-1 training. The center line and shaded region indicate the mean and standard deviation across 5 versions. (c) and (d) show the percentage of the samples that converges within 50 steps. (g) indicates the convergence speed.}
    \label{idct_A}
\end{figure}

Fig. \ref{idct_A} shows the effect of converting the observation vector to a high-dimensional space through inverse discrete cosine transform. First, a higher input dimensionality increases the number of parameters in the first weight matrix (the weight matrix closest to the input), which yields better convergence speed, as shown in Fig. \ref{idct_A} (g). 
Second, we define the step to convergence as follows. Keep monitoring the \(\mathrm{MaxStateChange} = \mathbf{max}(\Delta \xi_i)\,\,\forall i\) before and after each EP state update in relaxation, where \(i\) is the index of the neuron set of the entire network. If for the previous 5 consecutive steps, \(\mathrm{MaxStateChange} < 10^{-4}\), the current step is marked as the step to convergence. In Fig. \ref{idct_A} (a),(b),(e),(f), it can be observed that, with iDCT preprocessing, the free phase of the policy network, the free and nudge phases of the value network requires a smaller number of steps to converge.
Besides, Fig. \ref{idct_A} (c),(d) shows the percentage of samples that can converge within 50 relaxation steps. This percentage reduces faster when there is no iDCT preprocessing, which implies a long convergence process for a greater number of samples.

\newpage

\section{EP-PPO algorithm}
\label{sec:epppo}

\begin{algorithmic}[1]
\Require policy network \(\mathbf{Policy}(s)\), value network \(\mathbf{Value}(s)\), log-std vector \(\log(\sigma)\)
\Require number of environments \(N\), rollout length \(T\), discount factor \(\gamma\), GAE parameter \(\lambda\), policy learning rate \(\eta_{\mathrm{policy}}\), policy learning rate lower and upper bounds \(\eta_{\mathrm{policy,lower}}, \eta_{\mathrm{policy,upper}}\), policy learning rate adaptation coefficient \(\kappa\), value learning rate \(\eta_{\mathrm{value}}\), log-std learning rate \(\eta_{\mathrm{logstd}}\)
\Require number of PPO epochs \(K_\mathrm{epoch}\), mini-batch size \(|\mathcal{B}|\), target KL/early-stop KL/rollback KL \(\mathrm{KL}_{\mathrm{target}}/\mathrm{KL}_{\mathrm{stop}}/\mathrm{KL}_{\mathrm{rb}}\)
\Require EP nudge coefficient \(\beta_\mathrm{ep}\), policy free/nudge steps \(T^\pi_{\mathrm{free}},\,T^\pi_{+},\,T^\pi_{-}\), value free/nudge steps \(T^V_{\mathrm{free}},\,T^V_{+},\,T^V_{-}\)
\Require iDCT matrix \(W_{\mathrm{iDCT}}\)
\State Initialize policy network parameters, value network parameters, and log-std vector \(\log(\sigma)\)
\State Initialize SGD optimizer for policy network parameters with lr \(=\eta_{\mathrm{policy}}\), momentum \(=0.9\), weight decay \(=0\)
\State Initialize SGD optimizer for value network parameters with lr \(=\eta_{\mathrm{value}}\), momentum \(=0.9\), weight decay \(=0\)
\State Initialize Adam optimizer for \(\log(\sigma)\) with lr \(=\eta_{\mathrm{logstd}}\), betas \(=(0.9,0.999)\), eps \(=10^{-8}\), weight decay \(=0\)
\State Initialize raw observation running normalizer \(\mathrm{Normalizer_{raw}}\)
\State Initialize iDCT observation running normalizer \(\mathrm{Normalizer_{iDCT}}\)
\State Initialize \(N\) environments
\State Initialize \(\mathrm{SampleCounter} = 0\)
\While{\(\mathrm{SampleCounter} < \mathrm{MaxTrainingSamples}\)}
    \newline
    \Statex \textit{ --- Rollout collection --- }
    \For{\(t = 0\) to \(T-1\)}
        \For{\(n = 1\) to \(N\)}
            \State Observe \(s_{n,t}\)
            \State Sample \(\varepsilon_{n,t}\sim\mathcal{N}(0,I)\)
            \State Execute \(a_{n,t} = \mathbf{Policy}(s_{n,t}) + \sigma \odot \varepsilon_{n,t}\)
            \State Step environment and store data tuple into buffer
            \[
            (s_{n,t},\, a_{n,t},\, r_{n,t},\, \mathrm{log}(\pi_{\mathrm{rollout}}(a_{n,t}|s_{n,t})),\, d_{n,t},\, \mathrm{fall}_{n,t},\, \mathbf{Value}(s_{n,t}),\, \mathbf{Value}(s_{n,t+1}))
            \] \Comment{See Appendix~\ref{sec:EP_A}}
        \EndFor
    \EndFor
    \State \(\mathrm{SampleCounter} \mathrel{+}= N\cdot T\)
    \newline
    \Statex \textit{ --- Advantage and return computation --- }
        \State Compute estimated \(\bigl\{(\mathbf{Return}(s_{n,t}),\mathbf{Advantage}(s_{n,t}, a_{n,t}))\bigr\}_{n=1,t=1}^{N,T}\), noted as  \(\bigl\{(\hat{R}_{n,t},\hat{A}_{n,t})\bigr\}_{n=1,t=1}^{N,T}\) using GAE and data in the buffer.
        Here the input to GAE is \(\bigl\{(r_{n,t},\, d_{n,t},\, \mathrm{fall}_{n,t},\, \mathbf{Value}(s_{n,t}),\, \mathbf{Value}(s_{n,t+1})\bigr\}_{n=1,t=1}^{N,T}\).
    \State Flatten all \(N\times T\) samples into
    \[
    \mathcal{D} = \bigl\{(s_t,\, a_t,\, \log(\pi_{\mathrm{rollout}}(a_{t}|s_{t})),\, \hat{A}_t,\, \hat{R}_t)\bigr\}_{t=1}^{NT}
    \]
    \State Apply normalization and iDCT to observations:
    \[
        s_t \gets \mathrm{Normalizer_{raw}}(s_t)
    \]
    \[
    s_t \gets \mathrm{Normalizer_{iDCT}}\!\bigl(W_{\mathrm{iDCT}}\cdot s_t\bigr)
    \]
    \State Update running normalizers \(\mathrm{Normalizer_{raw}},\,\mathrm{Normalizer_{iDCT}}\)
    \State Normalize advantages:
    \[
    \hat{A}_t \gets \frac{\hat{A}_t - \mu_A}{\sigma_A + 10^{-8}},\quad \forall t
    \]
    where \(\mu_A, \sigma_A\) are the mean and standard deviation of the raw advantages over the rollout
    \State Snapshot policy network parameter and log-std vector\((\mathbf{Policy}_{\mathrm{snap}},\log(\sigma)_{\mathrm{snap}}) \gets (\mathbf{Policy},\log(\sigma))\), where \(\mathbf{Policy}_{\mathrm{snap}}\)
    \State caches old means \(\mu^{\mathrm{old}}_t \gets \mathbf{Policy}(s_t)\)
    \newline
    \Statex \textit{ --- Policy update with EP three-phase --- }
    \For{\(k=1\) to \(K_\mathrm{epoch}\)}
        \State Shuffle \(\mathcal{D}\)
        \For{each mini-batch \(\mathcal{B}\subset\mathcal{D}\)}
            \State \textbf{Free phase:} 
            \State Initialize policy neurons \(\xi\); 
            \State Relaxation for \(T^\pi_{\mathrm{free}}\) steps with \(\beta = 0\); 
            \State Obtain fixed point \(\xi^*\)
            \State \textbf{Positive nudge:} 
            \State Initialize policy neurons \(\xi \gets \xi^*\)
            \State Relaxation for \(T^\pi_{+}\) steps with loss gradient in Equation \ref{modified_policy_loss_grad} and \(\beta = \beta_\mathrm{ep}\)
            \State Obtain fixed point \(\xi^{+\beta_\mathrm{ep}}\)
            \State \textbf{Negative nudge:} 
            \State Initialize policy neurons \(\xi \gets \xi^*\)
            \State Relaxation for \(T^\pi_{-}\) steps with loss gradient in Equation \ref{modified_policy_loss_grad} and \(\beta = -\beta_\mathrm{ep}\)
            \State Obtain fixed point \(\xi^{-\beta_\mathrm{ep}}\) 
            \State Compute policy network parameter gradient by using \(\xi^{+\beta_\mathrm{ep}}, \xi^{-\beta_\mathrm{ep}}, \beta_\mathrm{ep}\) with Equation \ref{ep_grad} and SGD step
            \State Adam step on \(\log(\sigma)\) with gradient in Equation \ref{logstd_grad_A} \Comment{log-std vector update}
        \EndFor
        \State Estimate analytic mean KL between current and snapshot policy:
        \[
        \widehat{\mathrm{KL}} = \frac{1}{2|\mathcal{D}|}\sum_t \sum_i^{D_{\mathrm{action}}} \frac{\|\mathbf{Policy}(s_t)_i - \mu_{t,i}^{\mathrm{old}}\|^{2}}{\sigma_i^{2}}
        \]
        The change of \(\sigma\) is ignored in the KL-div calculation because \(\sigma\) only changes slightly for each rollout.
        \If{\(\widehat{\mathrm{KL}} > \mathrm{KL}_{\mathrm{stop}}\)}
            \State \textbf{break} \Comment{KL early stopping of policy epochs}
        \EndIf
    \EndFor
    \newline
    \Statex \textit{ --- Value update with EP three-phase --- }
    \For{\(k=1\) to \(K_\mathrm{epoch}\)}
        \State Shuffle \(\mathcal{D}\)
        \For{each mini-batch \(\mathcal{B}\subset\mathcal{D}\)}
            \State \textbf{Free phase:} 
            \State Initialize value neurons \(\xi_v\); 
            \State Relaxation for \(T^V_{\mathrm{free}}\) steps with \(\beta = 0\); 
            \State Obtain fixed point \(\xi_v^{*}\)
            \State \textbf{Positive nudge:} 
            \State Initialize value neurons \(\xi_v \gets \xi_v^{*}\)
            \State Relaxation for \(T^V_{+}\) steps with loss in Equation \ref{valueLoss_A} and \(\beta = \beta_\mathrm{ep}\)
            \State Obtain fixed point \(\xi_v^{+\beta_\mathrm{ep}}\)
            \State \textbf{Negative nudge:} 
            \State Initialize value neurons \(\xi_v \gets \xi_v^{*}\)
            \State Relaxation for \(T^V_{-}\) steps with loss in Equation \ref{valueLoss_A} and \(\beta = -\beta_\mathrm{ep}\)
            \State Obtain fixed point \(\xi_v^{-\beta_\mathrm{ep}}\) 
            \State Compute value network parameter gradient by using \(\xi_v^{+\beta_\mathrm{ep}}, \xi_v^{-\beta_\mathrm{ep}}, \beta_\mathrm{ep}\) with Equation \ref{ep_grad} and SGD step
        \EndFor
    \EndFor
    \newline
    \Statex \textit{ --- KL safeguard and adaptive LR --- }
    \If{\(\widehat{\mathrm{KL}} > \mathrm{KL}_{\mathrm{rb}}\)}
        \State Restore \((\mathbf{Policy},\,\log(\sigma)) \gets (\mathbf{Policy}_{\mathrm{snap}},\,\log(\sigma)_{\mathrm{snap}})\) \Comment{policy and log-std rollback}
        \State \(\eta_{\mathrm{policy}} \gets \eta_{\mathrm{policy}} / \kappa^2\)
    \ElsIf{\(\widehat{\mathrm{KL}} > 2\,\mathrm{KL}_{\mathrm{target}}\)}
        \State \(\eta_{\mathrm{policy}} \gets \eta_{\mathrm{policy}} / \kappa\)
    \ElsIf{\(\widehat{\mathrm{KL}} < 0.5\,\mathrm{KL}_{\mathrm{target}}\)}
        \State \(\eta_{\mathrm{policy}} \gets \kappa\,\eta_{\mathrm{policy}}\)
    \EndIf
    \State \(\eta_{\mathrm{policy}} \gets \mathbf{Clamp}(\eta_{\mathrm{policy}}, \eta_{\mathrm{policy,lower}}, \eta_{\mathrm{policy,upper}})\) \Comment{Clip policy learning rate}
\EndWhile
\end{algorithmic}

\newpage

\section{List of hyper-parameters}
\label{sec:hyper_parameter_list}

\begin{table}[h]
\renewcommand{\arraystretch}{1.3}
\centering
\caption{List of hyper-parameters. If a hyper-parameter is different in stage-1 and stage-2, the value in each stage will be displayed individually in the form (\textbf{1:} value1, \textbf{2:} value2).}
\label{list_hyper_parameter}
\begin{tabular}{@{} >{\raggedright\arraybackslash}p{3cm} >{\centering\arraybackslash}p{3cm} >{\arraybackslash}p{7cm} @{}}
\toprule
\textbf{Hyper-parameter} & \textbf{Value} & \textbf{Description} \\
\midrule

\(N\) & (\textbf{1:} \(1024\), \textbf{2:} \(2048\)) & Number of simulation environments\\

- & \(2000\) & Maximum episode length\\

\(v_{x,\mathrm{upper}}\) & \(\qty{0.5}{\meter/\second}\) & Upper bound of local x-directional target velocity\\

\(\mathbf{ResLimit}\) & \(1.0\) rad & Residual angle integration bound\\

\(\mathbf{TorqueLimit}\) & \(\qty{33.5}{\newton\meter}\) & Robot actuator torque bound\\

\(\mathbf{SideLength_{low}}\) & \(\qty{0.3}{\meter}\)& Terrain box side length lower bound\\

\(\mathbf{SideLength_{high}}\) & \(\qty{0.5}{\meter}\)& Terrain box side length upper bound\\

\(\mathrm{MaxTrainingSamples}\) & \(10^8\) & Total training samples\\

\(\gamma\) & \(0.99\) & RL discount factor\\

\(\lambda\) & \(0.95\) & GAE lambda\\

\(T\) & (\textbf{1:} \(256\), \textbf{2:} \(128\)) & Per-env rollout length\\

\(K_{\mathrm{epoch}}\) & \(10\) & Number of training epochs for policy and value networks\\

- & \(4\) & Number of training mini-batches in each epoch for policy and value networks\\

\(\mathrm{KL}_{\mathrm{target}}\) & \(0.01\) & Target KL-divergence\\

\(\mathrm{KL}_{\mathrm{stop}}\) & \(0.02\) & Early-stop KL-divergence\\

\(\mathrm{KL}_{\mathrm{rb}}\) & \(0.04\) & Rollback KL-divergence\\

- & \(17.03\) & Entropy scheduler initial entropy\\

- & \(17.03\) & Entropy scheduler final entropy\\

\(k_{\mathrm{entropy}}\) & \(0.01\) & Entropy loss coefficient\\

\(\eta_{\mathrm{logstd}}\) & \(3\times10^{-4}\) & Constant log-std learning rate\\

\(\eta_{\mathrm{policy,initial}}\) & \(0.1\) & Initial policy learning rate\\

\(\eta_{\mathrm{policy,lower}}\) & \(10^{-6}\) & Policy learning rate lower bound\\

\(\eta_{\mathrm{policy,upper}}\) & \(10\) & Policy learning rate upper bound\\

\(\kappa\) & \(1.5\) & Policy learning rate adaptation coefficient\\

\(\eta_{\mathrm{value}}\) & \(0.1\) & Constant value learning rate\\

- & Float32 & EP network datatype for neuron state and model parameters\\

\(\epsilon\) & \(0.2\) & PPO ratio clip coefficient\\

\(\epsilon_{\mathrm{rev}}\) & \(0.7\)  & Reverse PPO ratio clip coefficient\\

\(\alpha_{w}\) & \(0.5\) & EP weight initialization scale\\

\(T^\pi_{\mathrm{free}},\,T^\pi_{+},\,T^\pi_{-}\) & \(30,20,10\) & Policy free/nudge steps\\

\(T^V_{\mathrm{free}},\,T^V_{+},\,T^V_{-}\) & \(25,15,10\) & Value free/nudge steps\\

\(\beta_{\mathrm{ep}}\) & \(0.1\) & EP beta\\

\(\epsilon_{\mathrm{ep}}\) & \(1.0\) & EP state update coefficient\\

\bottomrule
\end{tabular}
\end{table}

\end{document}